\documentclass[journal]{IEEEtran}
\usepackage{lineno,hyperref}
\usepackage{bm}
\usepackage{amsmath}
\usepackage{graphicx} 
\usepackage{grffile}

\usepackage{amssymb}
\usepackage{epstopdf}
\usepackage{booktabs}
\usepackage{threeparttable}
\usepackage{multicol}
\usepackage{multirow}
\usepackage{subfigure}
\usepackage{algorithm}
\usepackage{algorithmicx}
\usepackage{algpseudocode}

\usepackage{lineno,hyperref}
\usepackage{bm}
\usepackage{algorithm}
\usepackage{algorithmicx}
\usepackage{algpseudocode}
\usepackage{epstopdf}
\usepackage{xcolor}

\usepackage{float}
\usepackage{cite}

\usepackage{lineno,hyperref}

\ifCLASSINFOpdf
\else
\fi
\begin{document}

\title{A Centroid Auto-Fused Hierarchical \\ Fuzzy c-Means Clustering}

\author{Yunxia~Lin~and~Songcan~Chen
\IEEEcompsocitemizethanks{\IEEEcompsocthanksitem The authors are with College of Computer Science and Technology/College of Artificial Intelligence, Nanjing University of Aeronautics and Astronautics, Nanjing, 211106, China and also with MIIT Key Laboratory of Pattern Analysis and Machine Intelligence. Corresponding author is Songcan Chen. E-mail: \{linyx, s.chen\}@nuaa.edu.cn.}
\thanks{}}

\markboth{Journal of \LaTeX\ Class Files,~Vol.~14, No.~8, August~2015}%
{Shell \MakeLowercase{\textit{et al.}}: Bare Demo of IEEEtran.cls for IEEE Journals}

\maketitle

\begin{abstract}
Like k-means and Gaussian Mixture Model (GMM), fuzzy c-means (FCM) with soft partition has also become a popular clustering algorithm and still is extensively studied. However, these algorithms and their variants still suffer from some difficulties such as determination of the optimal number of clusters which is a key factor for clustering quality. A common approach for overcoming this difficulty is to use the trial-and-validation strategy, i.e., traversing every integer from large number like $\sqrt{n}$ to 2 until finding the optimal number corresponding to the peak value of some cluster validity index. But it is scarcely possible to naturally construct an adaptively agglomerative hierarchical cluster structure as using the trial-and-validation strategy. Even possible, existing different validity indices also lead to different number of clusters. To effectively mitigate the problems while motivated by convex clustering, in this paper we present a Centroid Auto-Fused Hierarchical Fuzzy c-means method (CAF-HFCM) whose optimization procedure can automatically agglomerate to form a cluster hierarchy, more importantly, yielding an optimal number of clusters without resorting to any validity index. Although a recently-proposed robust-learning fuzzy c-means (RL-FCM) can also automatically obtain the best number of clusters without the help of any validity index, so-involved 3 hyper-parameters need to adjust expensively, conversely, our CAF-HFCM involves just 1 hyper-parameter which makes the corresponding adjustment is relatively easier and more operational. Further, as an additional benefit from our optimization objective, the CAF-HFCM effectively reduces the sensitivity to the initialization of clustering performance. Moreover, our proposed CAF-HFCM method is able to be straightforwardly extended to various variants of FCM. Finally, extensive experiments on both synthetic and real data sets demonstrate the effectiveness and efficiency of CAF-HFCM.
\end{abstract}


\IEEEpeerreviewmaketitle

\section{Introduction}
\IEEEPARstart{L}{ike} k-means and Gaussian Mixture Model (GMM), fuzzy c-means (FCM) \cite{FCM:Bezdek} has also become a classical clustering algorithm and still is constantly studied so far \cite{FCM:Gu J, FCM:Z. Moslehi, FCM:T. Lei}. In contrast to hard clustering counterparts which force each sample to belong to only one class, as a soft clustering algorithm, FCM allows each sample to belong to multiple partitions with varying degrees of membership. A large body of extensions of FCM \cite{FCM:M.N. Ahmed,FCM:N.R. Pal,FCM:S.C. Chen}
 have been proposed and shown its superiority in capturing the cluster structure because of the additional flexibility.
\par{However, similar to those classical prototype-based clustering methods such as k-means \cite{prototypeBased:J.A. Hartigan}, Gaussian Mixture Model(GMM) \cite{prototypeBased:D.A. Reynolds}, FCM and its variants still suffer from some difficulties in determining the optimal number of clusters inherent in data which plays a key role for cluster quality and knowledge discovery. The various algorithms to overcome this difficulty mainly cover two categories: the one is explicitly resorting to cluster validity index and the other is implicitly determined in optimizing a predefined clustering objective function.
\par{For the first category, a common technique is to use trial-and-validation strategy, i.e., traversing a large number like $\sqrt N$ to 2 until finding the number corresponding to the optimal value of some validity index. In practice, there are many cluster validity indices, e.g., Dunn index \cite{ClusterValidity:J.C. Dunn}, Calinski–Harabasz (CH) index \cite{ClusterValidity:T. Calinski},
C index (CI)\cite{ClusterValidity:L.J. Hubert}, Davies–Bouldin index (DB) \cite{ClusterValidity:D.L. Davies}},
partition entropy (PE) \cite{ClusterFCMValidity:J.C. Bezdek2}, Xie–Beni index (XB) \cite{ClusterFCMValidity:X.L.Xie}, Akaike information criterion (AIC) \cite{ClusterFCMValidity:Bozdogan}, Bayesian information criterion (BIC) \cite{ClusterFCMValidity:Schwarz}, while different validity indices often lead to different cluster numbers. Moreover, the whole trial-and-validation process is computationally expensive because one has to perform the clustering on a wide range of predefined numbers.
\par{For the second category, the determination of the number of clusters is automatically fused into optimization procedure for a predefined clustering objective function. For example, by using a regularization parameter to control agglomeration, \cite{ClusterFCMNum:H. Frigui} proposed a robust competitive agglomerative (RCA) algorithm, which iteratively discards some cluster whose cardinality dropping below a threshold to update the number of clusters. By introducing the volume cluster prototype and redefining the distance between the data point and the cluster prototype, \cite{ClusterFCMNum:U. Kaymak} proposes the Extended-FCM (E-FCM). In each iteration of E-FCM, the two clusters whose similarity is higher than a threshold are merged until the objective function converges. By adding the negative object-to-clusters membership entropy as a regularization term to the fuzzy c-means objective function, \cite{ClusterFCMNum:M. Li} presents the agglomerative fuzzy c-means clustering, the algorithm gradually merges cluster centers by increasing the value of penalty factor of the negative entropy regularization term until the cluster number tends to be stable in consecutive iteration in seeking solution. Similarly, \cite{ClusterFCMNum:M.S. Yang} introduces alternative entropy-type terms to fuzzy c-means algorithm to put forward robust-learning FCM (RL-FCM). RL-FCM obtains the optimal cluster number by discarding clusters whose mixing proportions are lower than $\frac{1}{n}$ ($n$ is the number of data points) during the iteration process until convergence. By adding a regularization term to the FCM objective function based on a focal point chosen in advance, \cite{ClusterFCMNum:P. Fazendeiro} proposes FCM with focal point (FCM-FP), which performs successively via increasing the value of the regularization parameter until finding the optimal cluster number corresponding to a peak value of the validity measure. However, these algorithms also suffer from respective disadvantages, for example, cumbersome parameter selection \cite{ClusterFCMNum:H. Frigui}, validation index dependence \cite{ClusterFCMNum:M. Li, ClusterFCMNum:P. Fazendeiro}, similarity threshold dependence \cite{ClusterFCMNum:U. Kaymak}, parameter update without theoretical guarantee \cite{ClusterFCMNum:M.S. Yang}, focal point dependence \cite{ClusterFCMNum:P. Fazendeiro}. Moreover, a selection model proposed in \cite{ClusterFCMNum:Y. Wang}, as another categorical method, is reviewed, which combines multiple pairs of algorithms and indexes to find the optimal cluster number based on competitive comprehensive fuzzy evaluation. Although the selection model performs reliably, it suffers from cumbersome experiments.

\par{In this paper we present an alternative approach named Centroid Auto-Fused Hierarchical Fuzzy c-means (CAF-HFCM). In order to reduce sensitivity to initialization, similar to existing method \cite{ClusterFCMNum:R. Krishnapuram} in which each cluster is initially approximated by many prototypes, i.e., the initial partition has an over specified number of clusters, ${a}\sqrt{n}$ data points are randomly chosen as the initial centroids, where $a$ is a positive constant in $[1,3]$. The biggest difference between our method and the second categorical methods as mentioned above \cite{ClusterFCMNum:H. Frigui,ClusterFCMNum:M. Li,ClusterFCMNum:M.S. Yang} lies in the formation process of final partition, inspired by convex clustering which shrinks the cluster centroids towards fusion, the $\ell_2$ norm penalty between the cluster centroids, called a fused term, is added to FCM's objective to promote {automatic fusion} of these centroids, leading to the adaptive agglomeration of clusters. As the cluster agglomeration proceeds, the stable partition corresponding to a wide range of value of the penalty factor of the $\ell_2$ term is finally obtained, consequently, yielding the "optimal" number of clusters, this means that the CAF-HFCM is able to automatically determine the optimal number of clusters. It is the very point that endows CAF-HFCM with the capability of subtly avoiding the defects of the existing two categorical methods. Furthermore, a cluster hierarchy as a byproduct can also naturally be generated during the optimization with progressively increasing value of regularization parameter without needing to cumbersomely traverse each $K$ in a predefined range, thus providing us with clustering with different granularities and interpretability for data cluster structures to a certain extent.

\par{The main contributions of our work can be summarized as follows:
\begin{itemize}
\item CAF-HFCM can automatically determine the optimal number of clusters without resorting to any cluster validity index.
\item CAF-HFCM can naturally form a hierarchical partition tree by progressively increasing the penalty factor value of centroid fusion term, revealing the cluster relationship hidden in data, thus compensating lack of FCM and its variants. Moreover, for determining the optimal number of clusters from a predefined range $[1,{a}\sqrt n]$, CAF-HFCM only needs once initialization while FCM needs reinitialization for each value.
\item CAF-HFCM can empirically be observed to be less sensitive to initialization than FCM even the same setting of cluster number.
\item CAF-HFCM can straightforwardly be extended to the variants of FCM.
\item CAF-HFCM can naturally reduce down to convex clustering \cite{ClusterFCMCC:T. D. Hocking}\cite{ClusterFCMCC:F. Lindsten} when $K=n$.

\end{itemize}

\par{ In the rest of this paper, Section 2 briefly overviews the standard FCM and robust-learning FCM. Section 3 proposes our Centroid Auto-Fused Hierarchical FCM algorithm (CAF-HFCM) in detail. Section 4 reports extensive experimental results and analysis. Finally, Section 5 concludes this paper with future research directions.

\section{Related Work}
Before developing our Centroid Auto-Fused Hierarchical FCM, we first briefly introduce related works including standard FCM and robust-learning FCM in this section.
\subsection{Fuzzy c-means Clustering}
Assuming each data point belongs to every cluster with a corresponding fuzzy membership, the standard objective function of FCM for grouping {$n$} data points into {$K$} clusters comes in the form:
\begin{equation}
J_{fuz} = \sum_{j=1}^{K}\sum_{i=1}^{n} {\mu_{ij}^b}||x_{i}-u_{j}||^2
\end{equation}
where $x_{i}\in R^d$ is the $i$th sample and $n$ is the total number of data points, respectively, $K$ is a predefined number of clusters, $u_{j}$ is the cluster centroid or prototype of the $j$th group, $\mu_{ij}$ instructs the fuzzy membership of the $i$th sample belonging to $j$th cluster, which is enforced to satisfy $\mu_{ij}\in[0,1], \ \sum_{j=1}^{{K}}\mu_{ij}=1$. Besides, the parameter $b$ is a weighting exponent to indicate the level of fuzziness. By zeroing the first derivatives of $J_{fuz}$ with respect to ${u_{j}}$ $(j = 1,\cdots, K)$ and $\mu_{ij}$ $(i = 1,\cdots, n; \ j = 1, \cdots, K)$ respectively, we can obtain two updated equations for $J_{fuz}$ to get its local optimum via iterative strategy:
\begin{equation}
u_{j} = \frac{\sum_{i=1}^{n}{\mu_{ij}^b}x_{i}}{\sum_{i=1}^{n}{\mu_{ij}^b}}
\end{equation}
\begin{equation}
\mu_{ij} = \frac{{(1/d_{ij})}^{1/(b-1)}}{\sum_{j=1}^{K}{(1/d_{ij})}^{1/(b-1)}},\ d_{ij}=||x_{i}-u_{j}||^2
\end{equation}
\par{Apparently, FCM performs flexible clustering because it allows samples to belong to multiple clusters with varying degrees of membership. However, FCM {also has} the following weaknesses:
\begin{itemize}
\item The number of clusters $K$ must be given in advance and for each such $K$, FCM has to re-initialize the centroids.
\item FCM is empirically more sensitive to the initialization than {that of} our CAF-HFCM.
\end{itemize}

\subsection{Robust-learning Fuzzy c-means Clustering}
Aiming to automatically obtain the best number of clusters and enforce robustness to initializations with free of the fuzziness index and parameter selection for FCM, the robust-learning fuzzy c-means (RL-FCM) is proposed in \cite{ClusterFCMNum:M.S. Yang}, which adds several entropy terms to FCM's objective. First, to endue freedom of the fuzziness index $b$ for FCM, the entropy term of memberships, $\sum_{j=1}^{c}\sum_{i=1}^{n} {\mu_{ij}\ln\mu_{ij}}$, is added, {where $c$ is the number of clusters}. Next, the mixing proportion $\alpha = (\alpha_{1},\cdots,\alpha_{c})$ of clusters is introduced, where $\alpha_{j}$ indicates the probability of one data sample belonging to the {$j$th} partition. Inherently, $-\ln \alpha_{j}$ represents the information for the occurrence of a data point belonging to the $j$th cluster. The entropy term $\sum_{j=1}^{c}\sum_{i=1}^{n} {\mu_{ij}\ln\alpha_{j}}$ is thereby utilized to summarize the average of information where a data point belongs to the corresponding partition over fuzzy memberships. Moreover, the entropy term $\sum_{j=1}^{c} {\alpha_{j}\ln\alpha_{j}}$ is used to represent the average of information for the occurrence of each data point belonging to the corresponding cluster. Therefore, the whole RL-FCM objective function is constructed as follows:

\begin{equation}
\begin{aligned}
J(\mathbf{P},\alpha,\mathbf{U}) = &\sum_{j=1}^{c}\sum_{i=1}^{n} {\mu_{ij}^b}||x_{i}-u_{i}||^2 -{r_{1}}\sum_{j=1}^{c}\sum_{i=1}^{n} {\mu_{ij}\ln\alpha_{j}}\\
               & +{r_{2}}\sum_{j=1}^{c}\sum_{i=1}^{n} {\mu_{ij}\ln\mu_{ij}}-{r_{3}n}\sum_{j=1}^{c}\alpha_{j}\ln\alpha_{j}\\
\text{s.t.}\qquad
\mu_{ij} \in & [0,1], \qquad i = 1,\cdots,n;\ j = 1,\cdots,c\\
\sum_{j=1}^{c}\mu_{ij}& = 1,  \qquad i = 1,\cdots,n\\
\alpha_{j} \in & [0,1], \qquad j = 1,\cdots,c\\
\sum_{j=1}^{c}\alpha_{j} & = 1,  \qquad j = 1,\cdots,c.\\
\end{aligned}
\label{RL-FCM}
\end{equation}
where $\mathbf{P} = [\mu_{ij}]_{i=1:n}^{j=1:{{c}}}$, $\mathbf{U} =  [u_j]_{j=1}^{{c}}$ define the membership matrix and the center matrix respectively, $r_{1},\ r_{2}, \ r_{3}  \geq 0$ are utilized to adjust bias.
\par{Obviously, RL-FCM can also automatically determine the cluster number without resorting to any validity index, but as can be seen from (\ref{RL-FCM})}, the cluster hierarchy {cannot} naturally be formed during the optimization procedure, while so-involved three parameters make its optimization more complicated than ours where just one regularization factor is involved.

\section{Centroid Auto-Fused Hierarchical FCM}
In this section, our Centroid Auto-Fused Hierarchical FCM (CAF-HFCM) algorithm is detailed. In the following, we {provide} its model formulation, solution as well as analysis for convergence in separated sub-sections respectively.
  \subsection{Model Formulation}

  As in RL-FCM, we start by randomly choosing a relatively large number of data points as the initial centroids with aiming to decrease sensitivity to initialization {to a certain extent}. Inspired by the convex clustering \cite{ClusterFCMCC:T. D. Hocking}\cite{ClusterFCMCC:F. Lindsten} which shrinks the centroids, a $\ell_2$ norm penalty between the cluster centroids, named a fused term, is then appended to FCM's objective to encourage the {automatic fusion} of these cluster centroids. To this end, we formulate the optimization problem of CAF-HFCM with the fixed fuzziness index $b = 2$ as follows:

  \begin{equation}
  \begin{aligned}
  & \underset{{\mu}_{ij},u_{j}}{\text{min}}
  & & \frac{1}{2}\sum_{i=1}^{n}\sum_{j=1}^{K}{\mu_{ij}^2}||x_{i}-u_{j}||^2+\gamma\sum_{k=1}^{K-1}\sum_{l=k+1}^{K}{||u_{k}-u_{l}||}_{2} \\
  & \text{s.t.}
  & & \mu_{ij} \in [0,1], \qquad i = 1,\cdots,n;\ j = 1,\cdots,K\\
  &&& \sum_{j=1}^{K}\mu_{ij} = 1, \qquad i = 1,\cdots,n.
  \end{aligned}
  \label{ConvFCM}
  \end{equation}
where $\gamma$ as a unique regularized parameter is incorporated to balance the model fit and the centroid coalescence {term}, $u_j$ corresponds to the $j$th cluster centroid. Here, we set $ K = \lfloor {a}\sqrt{n} \rfloor$ with ${a}\in[1,3]$, where $\lfloor x \rfloor$ defines rounding down operation on $x$.

  \par{
  When $\gamma = 0$, obviously, (\ref{ConvFCM}) is reduced to the original FCM, and $n$ instances are partitioned into $K$ clusters with cluster overlap. As $\gamma$ progressively increases, the centroids obtained corresponding to the previous $\gamma$ are taken as the initial ones for the next $\gamma$, in this way, the initial centroids begin to being automatically fused, urging the clusters to agglomerate progressively, making the number of clusters {decrease} gradually during successive optimization procedure of CAF-HFCM with increasing value of $\gamma$. As the cluster agglomeration proceeds, a stable partition corresponding to a wide range of the regularized parameter is finally achieved, consequently, yielding the "optimal" number of clusters. As $\gamma$ is sufficiently large, all centroids are fused to the same one, also {enforcing all data points to coalesce into a single cluster}. Obviously, for determining the optimal cluster number from the range of $[1,{a}\sqrt n]$, CAF-HFCM only needs once initialization. Furthermore, a cluster hierarchy as an extra reward can also be naturally generated during the optimization with progressively increasing value of regularization parameter, concrete implementation is in Algorithm \ref{CAF-HFCM}.

  \par{When taking $K = n$, our presented CAF-HFCM reduces to the convex clustering with $w_{ij}=1$, where $w_{ij}$ is a positive weight exerting on the auto-fused term of centroid $i$ and centroid $j$.}
 {Here, we place emphasis on the discussion of the automatic nature of our proposed algorithm: we randomly choose $a\sqrt n$ data points as the initial cluster centroids, from (\ref{ConvFCM}), we can see that as the value of the penalty factor of fusion term is increased, these initial cluster centroids are progressively and automatically merged. At the beginning of the optimization, we set $\gamma$ as a small number, some initial centroids can be easily fused into the same locations. As $\gamma$ is iteratively updated by $\gamma = \gamma + \epsilon$, when $\gamma$ arrives at a specific value, i.e., the cluster number exactly corresponds to the possibly optimal one, continuously adding the increment $\epsilon$ can hardly change the number of already generated clusters. On the other hand, we can find that the so-found cluster number is not quite sensitive to $\gamma$  within a wide range. Once the "optimal" cluster number is obtained, if still wanting the cluster centroids to form a whole hierarchy of clusters, we can continuously increase the value of $\epsilon$, equivalently, $\gamma$, implying that when $\gamma$ is increased to a sufficient large value, all the clusters' centroids will be coalesced into a desirable hierarchy with single root. Therefore, the optimal cluster number is automatically determined and a hierarchy of clusters as a byproduct is naturally formed in the whole optimization process.}

  \subsection{Problem Solution}
  Let $\mathbf{U} =  [u_j]_{j=1}^K$ represent the centroid matrix where $u_j$ is the $j$th column of $\mathbf{U}$ and $\mathbf{P} = [\mu_{ij}]_{i=1:n}^{j=1:K}$ denote the membership matrix. The objective function in (\ref{ConvFCM}) is not convex for all variables $\mathbf{P}$, $\mathbf{U}$, simultaneously. To solve this optimization problem, we take into account an alternating procedure that alternates between membership matrix $\mathbf{P}$ and centroid matrix $\mathbf{U}$. Specifically, we perform optimization according to {the next three steps}.
  \par{Step 1: Solving $\mathbf{P}$ for a fixed $\mathbf{U}$}

  \par{To update $\mathbf{P}$ for a fixed $\mathbf{U}$, we need to minimize the function:

  \begin{equation}
  \begin{aligned}
  f(\mathbf{P}) = & \frac{1}{2}\sum_{i=1}^{n}\sum_{j=1}^{K}{\mu_{ij}^2}||x_{i}-u_{j}||^2\\
  \text{s.t.} \qquad
  & \mu_{ij} \in [0,1], \qquad i = 1,\cdots,n;\ j = 1,\cdots,K\\
  & \sum_{j=1}^{K}\mu_{ij} = 1, \qquad i = 1,\cdots,n.
  \end{aligned}
  \label{CAF-HFCMFormula}
  \end{equation}
  Setting the derivative of $f(\mathbf{P})$ w.r.t. $\mathbf{P}$ to zero, we have
  \begin{equation}
  \mu_{ij} = \frac{{1/d_{ij}}}{\sum_{i=1}^{{K}}{(1/d_{ij})}},\ d_{ij}=||x_{i}-u_{j}||^2
  \label{fP}
  \end{equation}
  Obviously, for updating $\mathbf{P}$, CAF-HFCM performs in the same way as FCM.

  \par{Step 2: Solving $\mathbf{U}$ for a fixed $\mathbf{P}$}

  \par{With $\mathbf{P}$ fixed, for obtaining $\mathbf{U}$, we need to minimize the objective function:
  \begin{equation}
  f(\mathbf{U}) = \frac{1}{2}\sum_{i=1}^{n}\sum_{j=1}^{K}{\mu_{ij}^2}||x_{i}-u_{j}||^2+\gamma\sum_{k<l}{||u_{k}-u_{l}||}_{2}
  \label{fU}
  \end{equation}
  {where} {$\sum_{k<l}=\sum_{k=1}^{K-1}\sum_{l=k+1}^{K}$.} To update $\mathbf{U}$, we resort to the alternating direction method of multipliers (ADMM) strategy proposed in \cite{CC:E.C. ChiA}. Let $v_{kl} = v_{k}-v_{l}$, (\ref{fU}) can be equivalently written as the following constrained problem:
  \begin{equation}
  \begin{aligned}
  & \underset{{u}_{j},v_{kl}}{\text{min}}
  & & \frac{1}{2}\sum_{i=1}^{n}\sum_{j=1}^{K}{\mu_{ij}^2}||x_{i}-u_{j}||^2+\gamma\sum_{k<l}{||v_{kl}||}_{2} \\
  & \text{s.t.}
  & & u_{k} - u_{l} - v_{kl} = 0, {k = 1,\cdots,K-1; l = 2,\cdots,K}.
  \end{aligned}
  \label{fUVnu}
  \end{equation}
  The augmented Lagrangian for (\ref{fUVnu}) is then given by
  \begin{equation}
  \begin{split}
  L_{\beta}(\mathbf{U,V,\Lambda})=\frac{1}{2}\sum_{i=1}^{n}\sum_{j=1}^{K}{\mu_{ij}^2}||x_{i}-u_{j}||^2+\gamma\sum_{k<l}{||v_{kl}||}_{2}\\
  +\sum_{k<l}\langle \mathbf{\lambda}_{kl},v_{kl}-u_{k}+u_{l}\rangle + \frac{\beta}{2}\sum_{k<l}||v_{kl}-u_{k}+u_{l}||_{2}^{2}
  \end{split}
  \label{Lagrange}
  \end{equation}
where $\mathbf\Lambda$ = $[\lambda_{kl}]$ is the Lagrange multiplier matrix, $\mathbf{V} = [v_{kl}]$ denotes the centroid difference matrix and $\beta$ is a nonnegative tuning parameter.
Fixing $\mathbf{V}$ and $\mathbf{\Lambda}$ to update $\mathbf{U}$, we need to minimize the function:
  \begin{equation}
  f(\mathbf{U}) = \frac{1}{2}\sum_{i=1}^{n}\sum_{j=1}^{K}{\mu_{ij}^2}||x_{i}-u_{j}||^2+\frac{\beta}{2}\sum_{k<l}||\tilde{v}_{kl}-u_{k}+u_{l}]||_{2}^{2}
  \label{updateU}
  \end{equation}
where $\tilde{v}_{kl}=v_{kl}+\frac{1}{\beta}\mathbf{\lambda}_{kl}$. Specifically, taking the first derivative of $f(\mathbf{U})$ w.r.t. $\mathbf{U}$, and zeroing it, then the updating formula for the $j$th column of $\mathbf{U}$ is as follows:
  \begin{equation}
  u_{j} = \frac{\sum_{i=1}^{n}{\mu_{ij}^2}x_{i}+\beta \alpha_{j}}
  {\sum_{i=1}^{n}{\mu_{ij}^2}+(n-1)u_{j}}
  \end{equation}
  where $\alpha_{j} = \sum_{l=j+1}^{K}\tilde{v}_{jl}-\sum_{k=1}^{j-1}\tilde{v}_{kj}+\sum_{l=j+1}^{K}u_{l}+\sum_{k=1}^{j-1}u_{k}$.
  To determine the centroid difference matrix $\mathbf{V}$ with fixed $\mathbf{U}$ and $\mathbf{\Lambda}$, the optimization problem (\ref{Lagrange}) is reformulated as follows:
  \begin{equation}
  \begin{aligned}
  f(\mathbf{V}) = &\gamma\sum_{k<l}||v_{kl}||_{2}+\sum_{k<l}\langle \mathbf{\lambda}_{kl},v_{kl}-u_{k}+u_{l}\rangle \\
                  & +\frac{\beta}{2}\sum_{k<l}||v_{kl}-u_{k}+u_{l}]||_{2}^{2}
  \end{aligned}
  \end{equation}
  Refering to \cite{CC:E.C. ChiA}, the $v_{kl}$ of $\mathbf{V}$ is determined by the proximal mapping
  \begin{equation}
  \begin{aligned}
  v_{kl} &= \mathop{\arg\min}_{v_{kl}} \frac{1}{2}[\|v_{kl}-(u_{k}-u_{l}-\frac{1}{\beta}{\mathbf{\lambda}_{kl})}\|_{2}^{2}+\frac{\gamma}{\beta}\|v_{kl}\|_{2}]\\
  & = prox_{\sigma_{kl}\|.\|}(u_{k}-u_{l}-\frac{1}{\beta}{\mathbf{\lambda}_{kl})},
  \end{aligned}
  \label{fV}
  \end{equation}
where $\sigma_{kl}=\frac{\gamma}{v_{kl}}$. To update the Lagrange multiplier matrix $\mathbf{\Lambda}$, we use the following formula:
  \begin{equation}
  \mathbf{\lambda}_{kl}=\mathbf{\lambda}_{kl}+\beta(v_{kl}-u_{k}-u_{l}).
  \label{flambda}
  \end{equation}
\par{Algorithm \ref{ADMM for U} (ADMM-U) summarizes the ADMM algorithm for updating $\mathbf{U}$.}

\par{Step 3: Merge equal elements of $\mathbf{U}$ according to the centroid difference matrix $\mathbf{V}$}
\par{We simply apply the breadth-first search to identify connected components of graph induced by {the centroid difference matrix} $\mathbf{V}$, i.e., to place an edge between $k$th and $l$th cluster centroids if $\mathbf{v}_{kl}=\mathbf{0}$, equivalent centroids in $\mathbf{U}$ are thereby merged and {then the number of centroids is updated}.}
\par{The entire optimization procedure for CAF-HFCM with fixed hyper-parameter is summarized in Algorithm \ref{SubAlgorithm}. The concrete implementation of CAF-HFCM is in Algorithm \ref{CAF-HFCM}, which clearly explains {the fact that} a cluster hierarchy can be naturally formed during the optimization with progressively increasing value of regularization parameter.

\begin{algorithm}
        \caption{ADMM-U}
        \footnotesize
        \begin{algorithmic}[1] 
            \Require data $\mathbf{X}\in R^{d\times n}$, membership matrix $\mathbf{P}$, initial number of clusters $\mathbf{K}$, hyper-parameter $\bm{\gamma}$.
            \Ensure The centroid matrix $\mathbf{U}$, The centroid difference matrix $\mathbf{V}$.
            \State $\mathbf{U}^{0} = \mathbf{U}^{m-1}$\ in step 4 of Algorithm \ref{SubAlgorithm}, initialize $\mathbf{V}^{0}$, $\mathbf{\Lambda}^{0}$ randomly.
            \For{each $m\in 1,2,3,\cdots $}
              \For{{each $j\in [1,n]$}}
                 \State Update ${\mathbf{U}}_{j}^{m}$ by taking ${\mathbf{P}}$, ${\mathbf{V}}^{m-1}$, ${\mathbf{\Lambda}}^{m-1}$ into Eq.(\ref{updateU})
              \EndFor
              \For {all k and l}
                 \State Update ${\mathbf{V}}_{kl}^{m}$ by taking ${\mathbf{U}}_{k}^{m}$, ${\mathbf{U}}_{l}^{m}$, ${\mathbf{\Lambda}}_{kl}^{m-1}$ into Eq.(\ref{fV})
                 \State Update ${\mathbf{\Lambda}}_{kl}^{m}$ by taking ${\mathbf{U}}_{k}^{m}$, ${\mathbf{U}}_{l}^{m}$, ${\mathbf{V}}_{kl}^{m}$, ${\mathbf{\Lambda}}_{kl}^{m-1}$ into Eq.(\ref{flambda})
              \EndFor
                 \If {$f(\mathbf{U})$ converges}
                 \State Break
                 \EndIf
            \EndFor
            \State
            \Return The centroid matrix $\mathbf{U}$, The centroid difference matrix $\mathbf{V}$.
        \end{algorithmic}
        \label{ADMM for U}
\end{algorithm}

\begin{algorithm}
        \caption{Centroid Auto-Fused Hierarchical FCM with fixed hyper-parameter}
        \footnotesize
        \begin{algorithmic}[1] 
            \Require data $\mathbf{X}\in \Re^{d\times n}$, initial number of clusters $\mathbf{K}$, hyper-parameter $\mathbf{\gamma}$.
            \Ensure Clustering struct $\bm{sol}$: The membership matrix $\mathbf{P}$, The centroid matrix $\mathbf{U}$, The number of clusters $\mathbf{c}$.
            \For{each $m\in 1,2,3,\cdots$}
              \For{{each {{$i \in [1,n]$}} \textbf{and} {each $j \in [1,$TempK$]$}}}
              \State Update ${\mathbf{P}}_{ij}^{m}$ by taking ${\mathbf{U}}_{j}^{m-1}$ into Eq.(\ref{fP})
              \EndFor
              \State Update ${\mathbf{U}}^{m}$ using Algorithm \ref{ADMM for U}
              \State Merge equal elements of ${\mathbf{U}}^{m}$ based on $\mathbf{V}$ obtained from the above step and update number of centroids TempK
              \If{(\ref{ConvFCM}) converges}
              \State Break
              \EndIf
            \EndFor
            \State
            \Return $\bm{sol}$: $\mathbf{P}$, $\mathbf{U}$, $\bm{c}=$ TempK.
        \end{algorithmic}
        \label{SubAlgorithm}
\end{algorithm}

\begin{algorithm}
        \caption{Centroid Auto-Fused Hierarchical FCM}
        \footnotesize
        \label{CAF-HFCM}
        \begin{algorithmic}[1] 
            \Require data $\mathbf{X}$ $\in$ $R^{d\times n}$, initial number of clusters $\mathbf{K}$, starting $\bm{\gamma}>0$, $\epsilon >0$, length of sequence of $\{\bm{\gamma_{i}}\}$ $l$, vector of cluster numbers $\{\bm{c_{i}}\}_{i=1}^{l}$ $=[0, \cdots,0]$.
            \Ensure Hierarchical clustering table $\bm{sol}$: The set of membership matrices $\{\mathbf{P}\}$, the set of centroid matrices $\{\mathbf{U}\}$, the optimal number of clusters $\bm{c}$.
            \State Initialize TempK $=\mathbf{K}$, randomly choose $\mathbf{K}$ data points and concatenate them by column as ${\mathbf{U}_{0}}^{0}$.
            \For{each $i\in [1,l]$}
            \State Obtain $\mathbf{P}_{i}$, $\mathbf{U}_{i}$, $c_{i}$ using Algorithm \ref{SubAlgorithm}.
            \State $\bm{\gamma}$$\leftarrow$$\bm{\gamma + \epsilon}$
            \State $\mathbf{K}=c_{i-1}$,\ $\mathbf{U}_{i}^{0}=\mathbf{U}_{i-1}$
            \EndFor
            \State Set the number in $\{\bm{c_{i}}\}_{i=1}^{l}$ that remains unchanged for a wide range of $\gamma$ as the optimal number of clusters $\bm{c}$
            \State
            \Return $\bm{sol}$: $\{\mathbf{P}\}$, $\{\mathbf{U}\}$, $\bm{c}$.
        \end{algorithmic}
\end{algorithm}

  \subsection{Convergence Analysis}
   \par{Obviously, as shown in (\ref{CAF-HFCMFormula}), the objective function of CAF-HFCM is convex in $\mathbf{P}$ for fixed $\mathbf{U}$, and convex in $\mathbf{U}$ for fixed $\mathbf{P}$, so the objective function (\ref{CAF-HFCMFormula}) of CAF-HFCM is biconvex \cite{Biconvex:J. Gorski}. Thus, according to the biconvex optimization theory \cite{Biconvex:J. Gorski}, by obtaining the corresponding optimization solution for each subproblem, a locally optimal solution thereby must be able to be found by our proposed optimization algorithm. Note that for our problem, each iterative update for $\mathbf{P}$ has an analytic solution, while for $\mathbf{U}$, we just can utilize ADMM method to solve it.}

\section{Experiments and Analysis}
In this section, we conduct {numerous different} experiments to validate the effectiveness and efficiency of the CAF-HFCM algorithm.

\subsection{Optimal Cluster Number Determination and Clustering Performance Comparison}
  {In this subsection, we perform CAF-HFCM on three synthetic and three real data sets to demonstrate the ability of CAF-HFCM to determine the optimal cluster number. Moreover, as the proposed CAF-HFCM can straightforwardly be extended to various variants of FCM, to further validate the clustering performance of CAF-HFCM over the optimal cluster number, we compare it with the following three clustering methods: Robust-Learning fuzzy c-means clustering algorithm (RL-FCM)\cite{ClusterFCMNum:M.S. Yang} (a recently proposed variant of FCM for automatically obtaining the optimal number of clusters), FCM (a typical soft clustering method), k-means \cite{prototypeBased:J.A. Hartigan} (a classical hard clustering method). Here, rand index (RI) \cite{RI}, adjusted rand index (ARI) \cite{ARI}, as well as normalized mutual information (NMI) \cite{NMI} are utilized as performance criteria. Note that the larger the values of these indices are, the better the clustering performance is. Besides, each experiment over optimal cluster number is repeated 20 times and the averaged results and corresponding standard deviations are tabulated in Table \ref{ComparisonRI}.}

  \par{\textbf{GaussianMixture Data.} 300 data points of a 2D Gaussian mixture distribution with mixture component ratio $\alpha_{k}=\frac{1}{6}$ are created as what described as \textbf{Example 4} in \cite{ClusterFCMNum:M.S. Yang}. Figure \ref{NumberOfClustersShowGaussianModifyData} shows the number of partitions of GaussianMixture data {with respect to different values of $\gamma$.} {In determining the optimal number of clusters, we seek adaptively the possibly optimal number of clusters by updating $\gamma$ using the additive rule $\gamma = \gamma + \epsilon$. Combining with Figure \ref{NumberOfClustersShowGaussianModifyData}, we emphasize that how to determine the optimal cluster number by utilizing CAF-HFCM. When $\gamma$ is increased to a specific value, i.e., the number of generated clusters exactly corresponds to the possibly optimal one, continuously adding the increment $\epsilon$ hardly changes the number of already formed clusters. On the other hand, we can find that the so-found cluster number is not quite sensitive to $\gamma$ within a wide range, e.g., $\gamma = 0.55-4.20$, that is, the cluster number basically stays unchanged for a long process of iteration in the Algorithm 3. From Figure \ref{NumberOfClustersShowGaussianModifyData}, we can see that CAF-HFCM detects 6 clusters for a wide range of $\gamma$, which exactly matches with the true size of clusters. Once the "optimal" cluster number is found, if still wanting the cluster centroids to form a whole hierarchy of clusters, we can continuously increase the value of $\epsilon$, equivalently, $\gamma$, implying that when $\gamma$ is increased to a sufficient large value, all the clusters' centroids will be coalesced into a desirable hierarchy with single root.}
 The {averages} of clustering performance of our proposed method {over the optimal cluster number} ($\gamma=2.18$, \ $c=6$) {are} 0.9934, 0.9760, 0.9726, which are higher around 4.31\%, 14.13\%, 5.97\% than {that} obtained by k-means. Even if CAF-HFCM performs similarly with RL-FCM, it is worthy to point out that the performance of RL-FCM heavily depends on the update rule of {its so-involved three} hyperparameters. Compared with RL-FCM, our CAF-HFCM involves just {one} parameter which makes the corresponding adjustment much easier and more operational. Though RI of FCM is similar with the one obtained by CAF-HFCM, note that the cluster number must be given in advance for FCM. }

  \begin{figure}
  \centering
  \includegraphics[width=9cm]{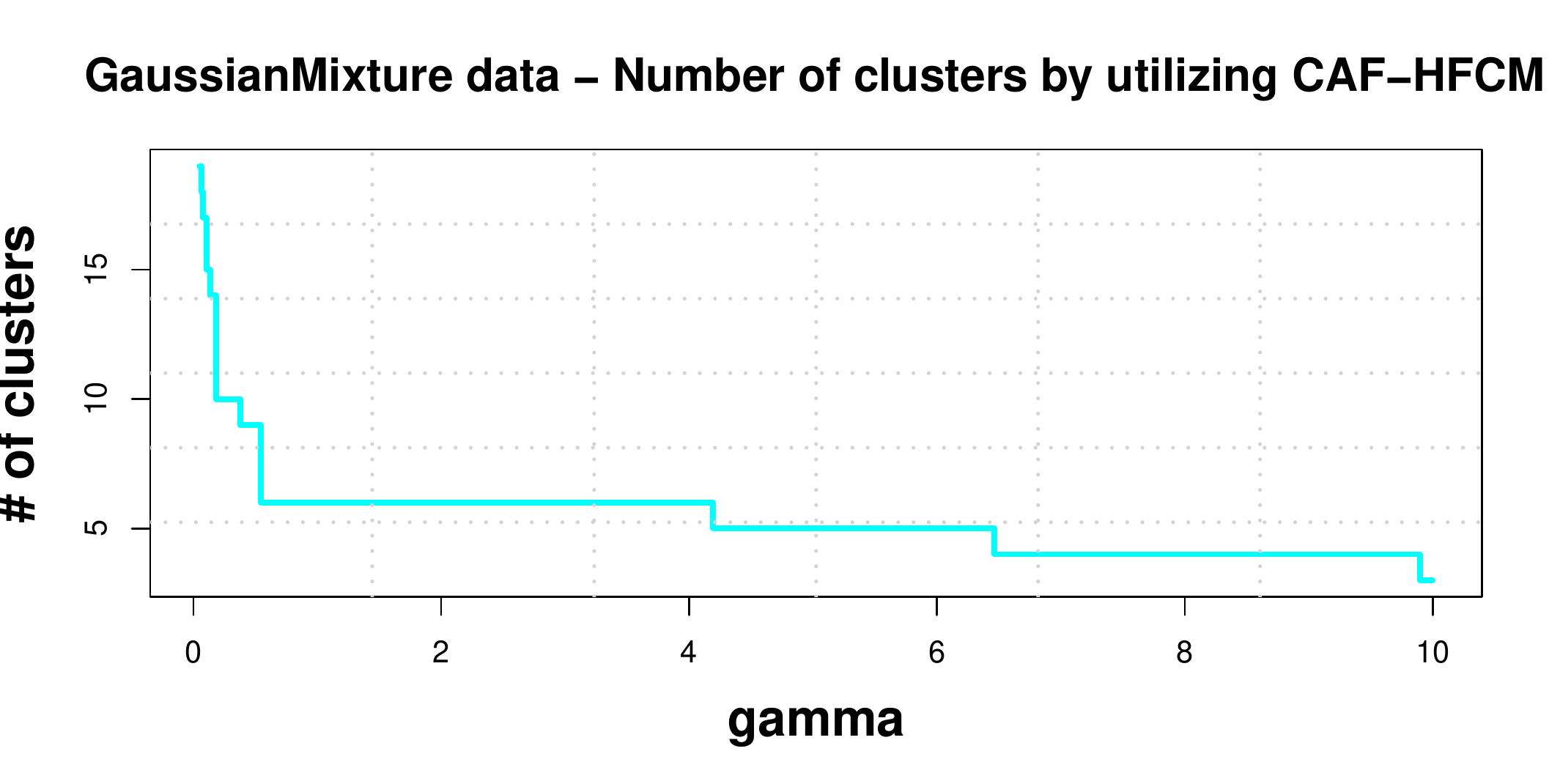}
  \caption{Number of clusters for GaussianMixture data under a sequence of $\gamma$.}
  \label{NumberOfClustersShowGaussianModifyData}
  \end{figure}

  \par{\textbf{GaussianMixture Data with High Dimension.} To test the performance of CAF-HFCM for dealing with data with high dimension, 300 20D Gaussian mixture data with $\alpha_{k} = \frac{1}{6}$ are generated. From Figure \ref{NumberOfClustersShow_HaussianMixture_High} we can see that the determined number of clusters is 6, which is rightly equal to the true cluster number. Setting $\gamma=9.25\ $ ($c=6$), as Table \ref{ComparisonRI} shows, the averages of clustering performance obtained by the proposed CAF-HFCM are RI = 1.0000, ARI = 1.0000, NMI = 1.0000 over the optimal cluster number, which are highest among those of all compared methods. {Although RL-FCM performs similarly with our proposed method over the optimal cluster number, we have to point out that RL-FCM can only obtain the optimal cluster number under the optimal update rule of the parameters for the penalty terms, while it is cumbersome to find.}

  \begin{figure}
  \centering
  \includegraphics[width=9cm]{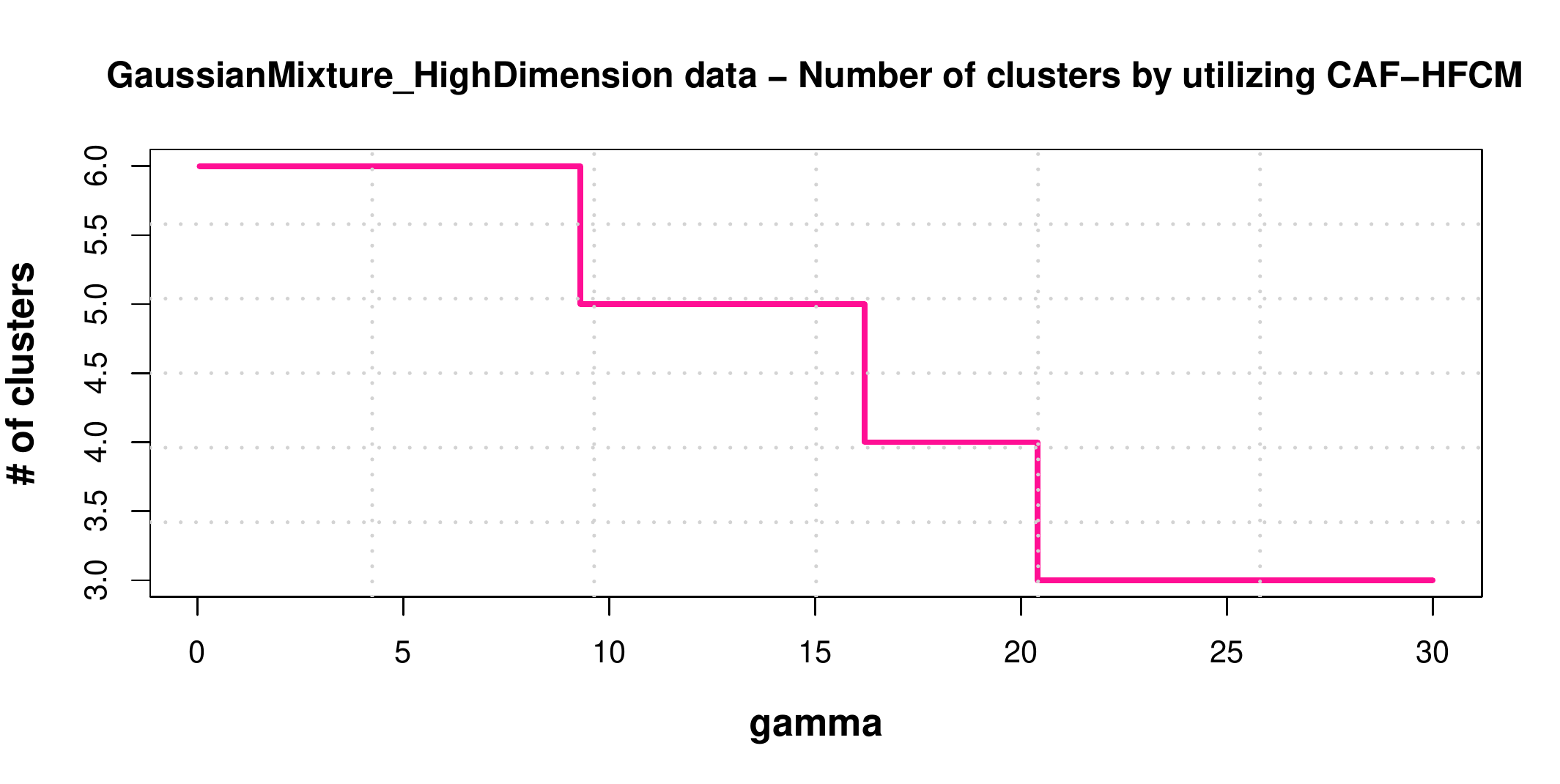}
  \caption{Number of clusters for GaussianMixture data with high dimension under a sequence of $\gamma$.}
  \label{NumberOfClustersShow_HaussianMixture_High}
  \end{figure}

  \par{\textbf{Gaussian Data.} In this synthetic example, we use a data set of 500 data points generated from 2 dimensional Gaussian distribution with similar covariance matrix $\Sigma=\begin{bmatrix}0.25&0\\0&0.25\end{bmatrix}$ and 25 different mean vectors $\mu = (1+10(i-1), \ 1+10(j-1))^{T}, (i = 1,\ \cdots,5; j = 1, \ \cdots, 4)$, as shown in Figure \ref{OriginalGaussianData}. Figure \ref{NumberOfClustersShowGaussianData} shows the number of partitions of Gaussian data controlled by a sequence of $\gamma$. From Figure \ref{NumberOfClustersShowGaussianData}, {we can see that when $\gamma$ arrives at a certain value, even continuously adding the increment $\epsilon$, the cluster number stays at 25 within a wide range of $\gamma$, so CAF-HFCM determines 25 as the optimal cluster number for Gaussian Data, which exactly coincides with the true one. Moreover, when setting $\gamma=0.25(c=25)$, the averages of RI, ARI, NMI of CAF-HFCM are 1.0000, 1.0000, 1.0000, respectively, which are the best among all the compared methods.}

 \begin{figure}
 \centering
 \includegraphics[width=7.5cm]{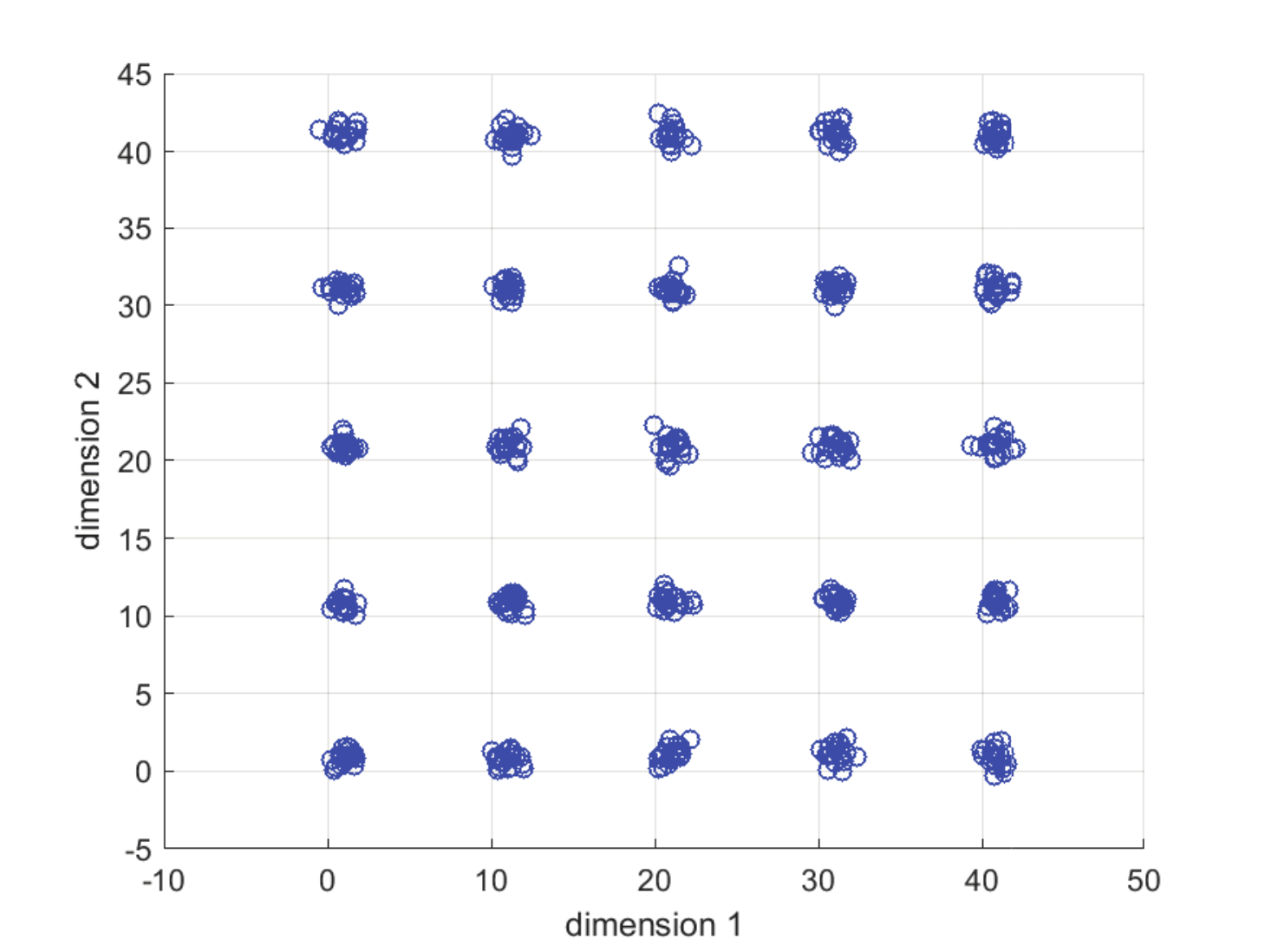}
 \caption{Original Gaussian data.}
 \label{OriginalGaussianData}
 \end{figure}

 \begin{figure}
 \centering
 \includegraphics[width=9cm]{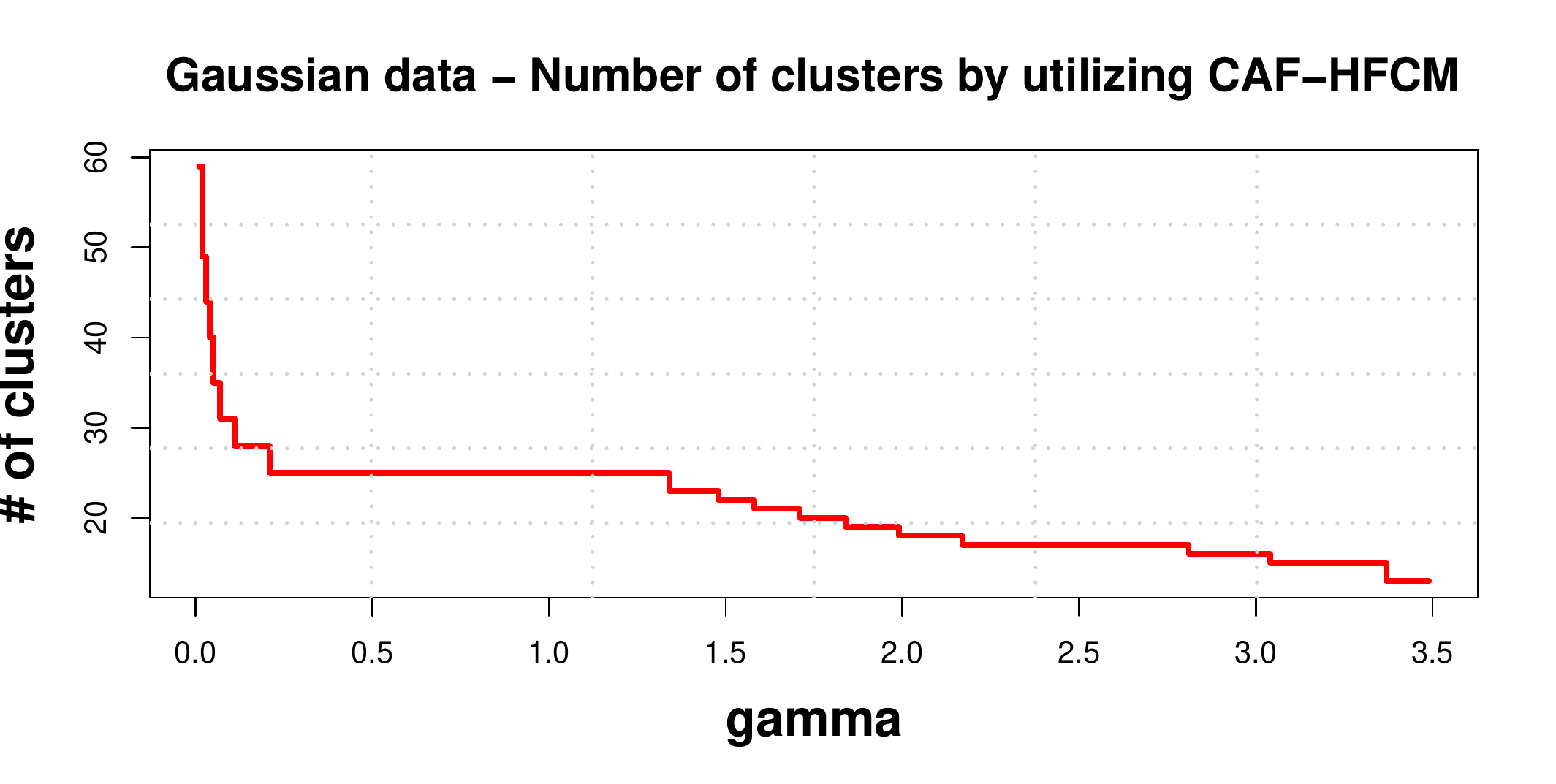}
 \caption{Number of clusters for Gaussian data under a sequence of $\gamma$.}
 \label{NumberOfClustersShowGaussianData}
 \end{figure}

  \textbf{Iris Data Set.} In this real example, we employ the Iris data set covering 150 data points with 4 attributes (i.e., sepal length(in cm), sepal width(in cm), petal length(in cm), and petal width (in cm)) from UCI machine Learning Repository \cite{DataSet:Iris}. The original Iris data set consists of 3 clusters whose name are setosa, versicolor, and virginica respectively. Figure \ref{NumberOfClustersShowIrisData} performs corresponding number of clusters of CAF-HFCM with respect to a sequence of $\gamma$. Referencing Figure \ref{NumberOfClustersShowIrisData}, {after $\gamma$ arrives at a certain value, the cluster number stays at 3 even if increasing its value continuously, therefore} we can {conclude} that {the} obtained number of clusters is $c = 3$, which {exactly} coincides with the true number of clusters. As $\gamma = 4.3 \ ($c = 3$)$, averages of respective RI, ARI, NMI obtained by CAF-HFCM are 0.9124, 0.8018, 0.7959 respectively, which are much higher 2.65$\%$, 5.73$\%$, 1.82$\%$ than that of RL-FCM and 3.27$\%$, 7.24$\%$, 4.63$\%$ than that of FCM. Actually, {note} that the cluster number of FCM must be set in advance, it is not capable of automatically obtaining the optimal number of clusters.

\begin{figure}
\centering
\includegraphics[width=9cm]{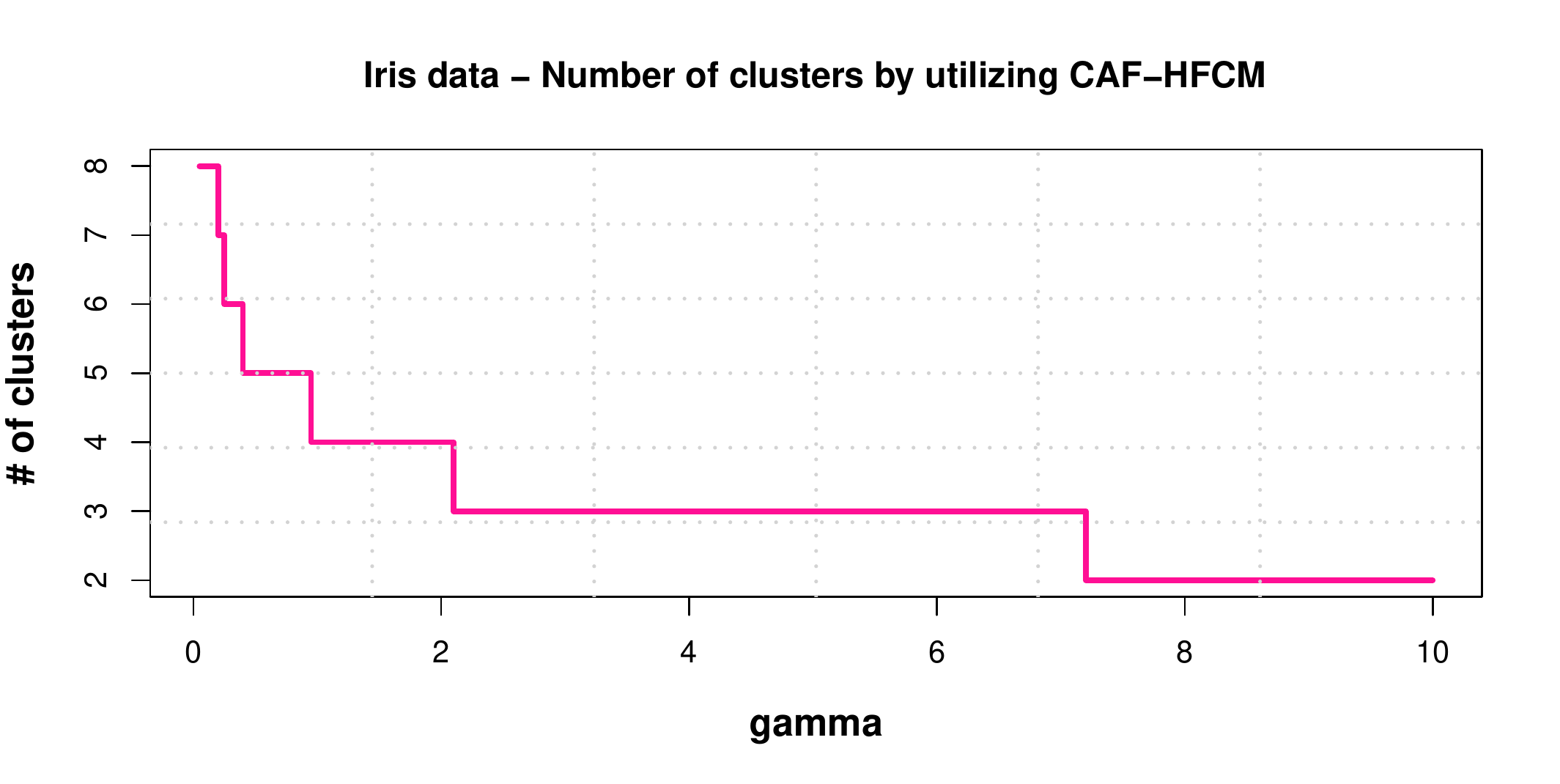}
\caption{Number of clusters for Iris data under a sequence of $\gamma$.}
\label{NumberOfClustersShowIrisData}
\end{figure}

\par{\textbf{Breast Data Set.} In this real example, the Breast data set from UCI machine Learning Repository \cite{DataSet:Iris} that contains 699 samples and 9 features as well as covers 2 clusters are employed. In this experiment, we discard one attribute with missing value. Figure \ref{NumberOfClustersShow_BreastData} performs number of clusters by utilizing CAF-HFCM with respect to a sequence of $\gamma$. And we can conclude that CAF-HFCM automatically {detects} 2 clusters for a wide range of $\gamma$, which exactly coincides with true number clusters. As setting $\gamma=100$ ($c = 2$), the averages of respective RI, ARI and NMI of cluster partitions determined by CAF-HFCM are 0.9151, 0.8284, 0.7231 respectively, which are higher around 0.52$\%$, 1.05$\%$, 1.35$\%$ than the performance of RL-FCM. {Although k-means performs similarly with CAF-HFCM, the number of clusters should be given in advance for k-means.}

\begin{figure}
\centering
\includegraphics[width=9cm]{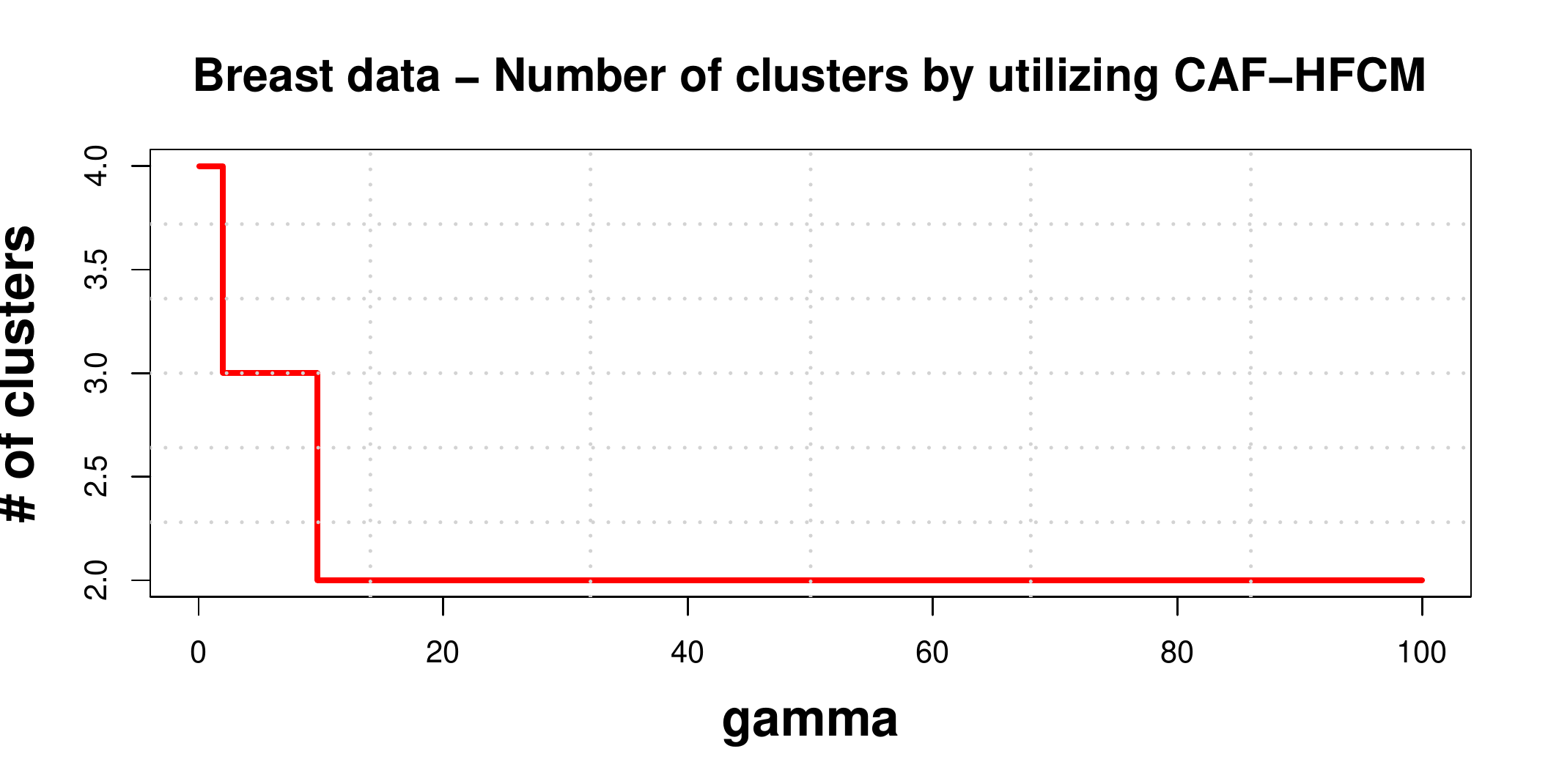}
\caption{Number of clusters for Breast data under a sequence of $\gamma$.}
\label{NumberOfClustersShow_BreastData}
\end{figure}

\par{\textbf{Seeds Data Set.} There exists 210 samples and 7 attributes in seeds data set \cite{DataSet:Iris}, which is partitioned into 3 clusters. Figure \ref{NumberOfClustersShow_SeedsData} describes the {determined} number of clusters {with respect to a sequence of hyperparameters $\gamma$}, and the corresponding obtained number exactly matches with true one. From Figure \ref{NumberOfClustersShow_SeedsData}, we find that CAF-HFCM obtains 3 clusters for a wide range of $\gamma$, this further demonstrates that CAF-HFCM can automatically determine the optimal cluster number. Setting $\gamma=14.5 \ $ ($c=3$), from Table \ref{ComparisonRI}, we can see that the proposed CAF-HFCM performs best with RI = 0.8814, ARI = 0.7331, NMI = 0.7229 among all the compared methods, which are higher around $0.7\%$, $1.65\%$, $2.8\%$ than that of {RL-FCM}. Actually, we should notice that for FCM and k-means, the {both} are not able to automatically determine optimal number of clusters, for RL-FCM, the one has so-involved three hyper-parameters to adjust.

{In summary, CAF-HFCM is not only capable of automatically determining the optimal cluster number without resorting to any validity index but also has the optimal clustering performance in most of cases. Furthermore, zero standard deviations demonstrate the lower sensitivity of CAF-HFCM to initialization than that of FCM. }

\begin{figure}
\centering
\includegraphics[width=9cm]{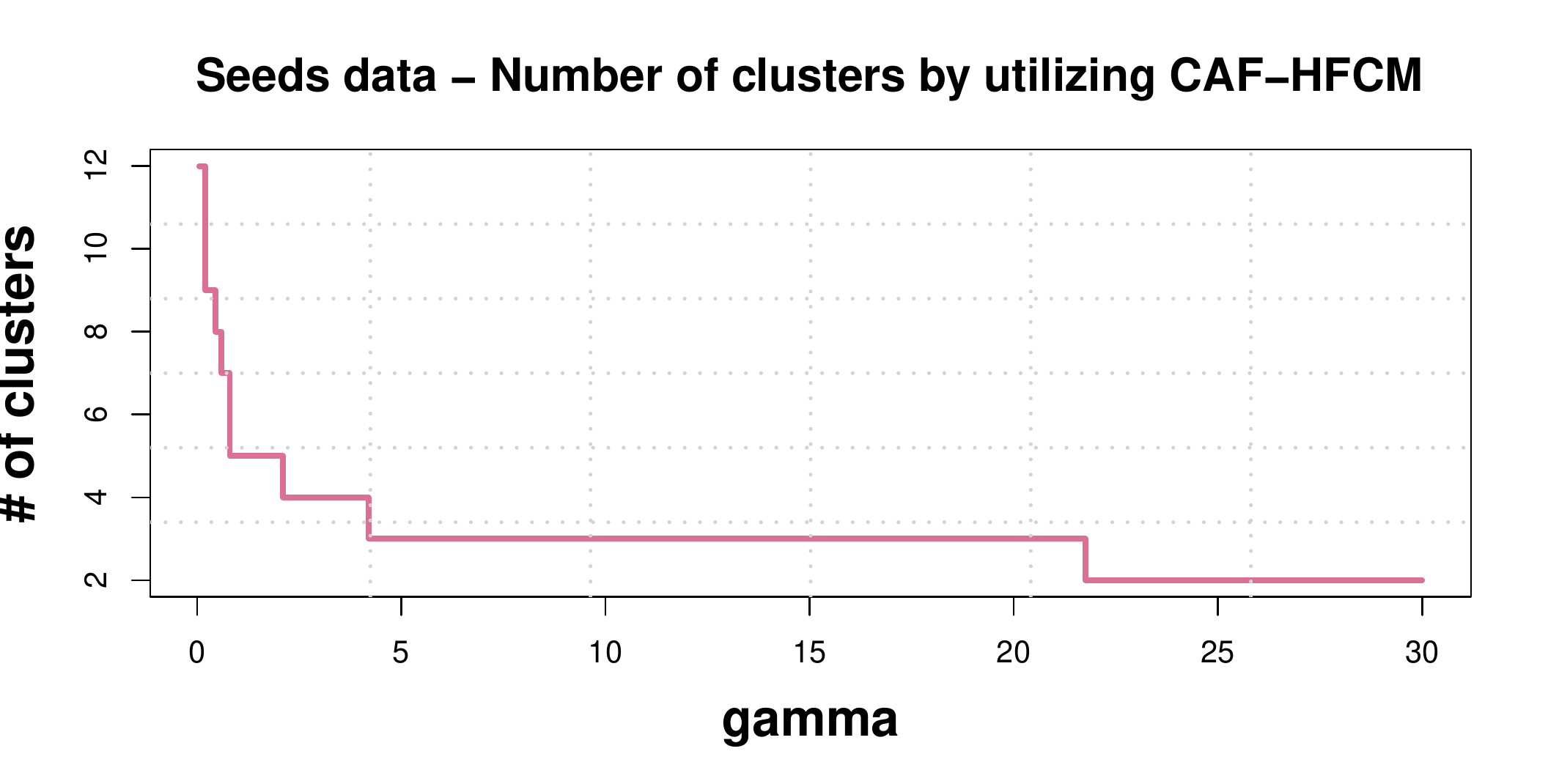}
\caption{Number of clusters for Seeds data under a sequence of $\lambda$.}
\label{NumberOfClustersShow_SeedsData}
\end{figure}

\begin{table*}[ht]
\renewcommand\arraystretch{1.2}
\tabcolsep 0pt
\caption{RI, ARI, and NMI from CAF-HFCM, RL-FCM, and FCM using the true number $c$ of clusters.}
\vspace*{-18pt}
\begin{center}
\def\temptablewidth{0.95\textwidth}
{\rule{\temptablewidth}{1pt}}
\begin{tabular*}{\temptablewidth}{@{\extracolsep{\fill}}cccccccccccc}
\hline

Dataset & Index & CAF-HFCM & RL-FCM & FCM & k-means
\\  \hline  

 \multirow{3}{*}{GaussianMixture} & RI & 0.9934$\pm$0 & 0.9934$\pm$0 & {\bf 0.9956$\pm$0.0155} &0.9503$\pm$0.0330\\ 
 & ARI & {\bf 0.9760$\pm$0}  & {\bf 0.9760$\pm$0} & { 0.9724$\pm$0.0512}  &0.8347$\pm$0.1081\\
 & NMI & {0.9726$\pm$0}  & {0.9726$\pm$0} & { \bf0.9765$\pm$0.0234} &0.9129$\pm$0.0471\\
 \hline

 \multirow{3}{*}{GaussianMixture-HD} & RI & {\bf 1.0000$\pm$0} &{\bf 1.0000$\pm$0} &  0.9870$\pm$0.0266 &0.9458$\pm$0.0311\\ 
 & ARI & {\bf 1.0000$\pm$0}  &{\bf 1.0000$\pm$0} & 0.9575$\pm$0.0873  &0.8249$\pm$0.0974\\
 & NMI & {\bf 1.0000$\pm$0}  &{\bf 1.0000$\pm$0}  & 0.9817$\pm$0.0377 &0.9242$\pm$0.0445\\
 \hline

 \multirow{3}{*}{Gaussian} & RI & {{\bf 1.0000$\pm$0}} & \bf 1.0000$\pm$0 & 0.9911$\pm$0.0035 & {0.9984$\pm$0.0023}\\ 
 & ARI & {{\bf 1.0000$\pm$0}}  &\bf 1.0000$\pm$0 & 0.8841$\pm$0.0434  & {0.9794$\pm$0.0307}\\
 & NMI & {{\bf 1.0000$\pm$0}}  &\bf 1.0000$\pm$0 & 0.9602$\pm$0.0159 & {0.9949$\pm$0.0076}\\
 \hline

 \multirow{3}{*}{Iris} & RI & {\bf 0.9124$\pm$0} &0.8859$\pm$0$\spadesuit$ & { 0.8797$\pm$0} &0.8687$\pm$0.0352\\ 
 & ARI & {\bf 0.8018$\pm$0}  &0.7445$\pm$0  & { 0.7294$\pm$0}  &0.7082$\pm$0.0661\\
 & NMI & {\bf 0.7959$\pm$0}  &0.7777$\pm$0  & { 0.7496$\pm$0} &0.7415$\pm$0.0372\\
 \hline

 \multirow{3}{*}{Breast} & RI & {0.9151$\pm$0} &{\ 0.9099$\pm$0} & 0.8996$\pm$0 &{\bf 0.9177$\pm$0}\\ 
 & ARI & {0.8284$\pm$0}  &{\ 0.8179$\pm$0}  & 0.7967$\pm$0  &{\bf 0.8338$\pm$0}\\
 & NMI & {0.7231$\pm$0}  &{\ 0.7096$\pm$0}  &0.6881$\pm$0 &{\bf 0.7289$\pm$0}\\
 \hline

 \multirow{3}{*}{Seeds} & RI & {\bf 0.8814$\pm$0 } & 0.8744$\pm$0 & 0.8744$\pm$0 & 0.8729$\pm$0.0015\\ 
 & ARI & {\bf 0.7331$\pm$0 }  & 0.7166$\pm$0 &  0.7166$\pm$0  &0.7135$\pm$0.0032\\
 & NMI & {\bf 0.7229$\pm$0 }  & 0.6949$\pm$0  & 0.6949$\pm$0 & 0.7025$\pm$0.0018\\
 \hline
\end{tabular*}
\label{ComparisonRI}
{\rule{\temptablewidth}{1pt}}

\emph{The results reflect independent 20 runs each algorithm over the optimal cluster number and expressed as mean $\pm$ standard deviation, in which the highest index value is with bold. $\spadesuit$ It is worthy to point out that the RI = $0.8859$ utilizing the code provided by authors of \cite{ClusterFCMNum:M.S. Yang}, which is different from the result RI = 0.8923 obtained in \cite{ClusterFCMNum:M.S. Yang}.}
\end{center}
\end{table*}

\subsection{Hierarchical Cluster Path Show}
In this subpart, for briefly showing the hierarchical clustering structure obtained by CAF-HFCM, 30 data points of a 2D Gaussian mixture distribution are constructed as seen in Figure \ref{GaussianMixture30}. Moreover, the corresponding hierarchical structure obtained by CAF-HFCM is shown in Figure \ref{GaussianMixture30ClusterPath}. As a byproduct, the hierarchical clustering path of the data set can provide us with clustering with different granularities and interpretability for data cluster structures to a certain extent.

\begin{figure}
\centering
\includegraphics[width=7.5cm]{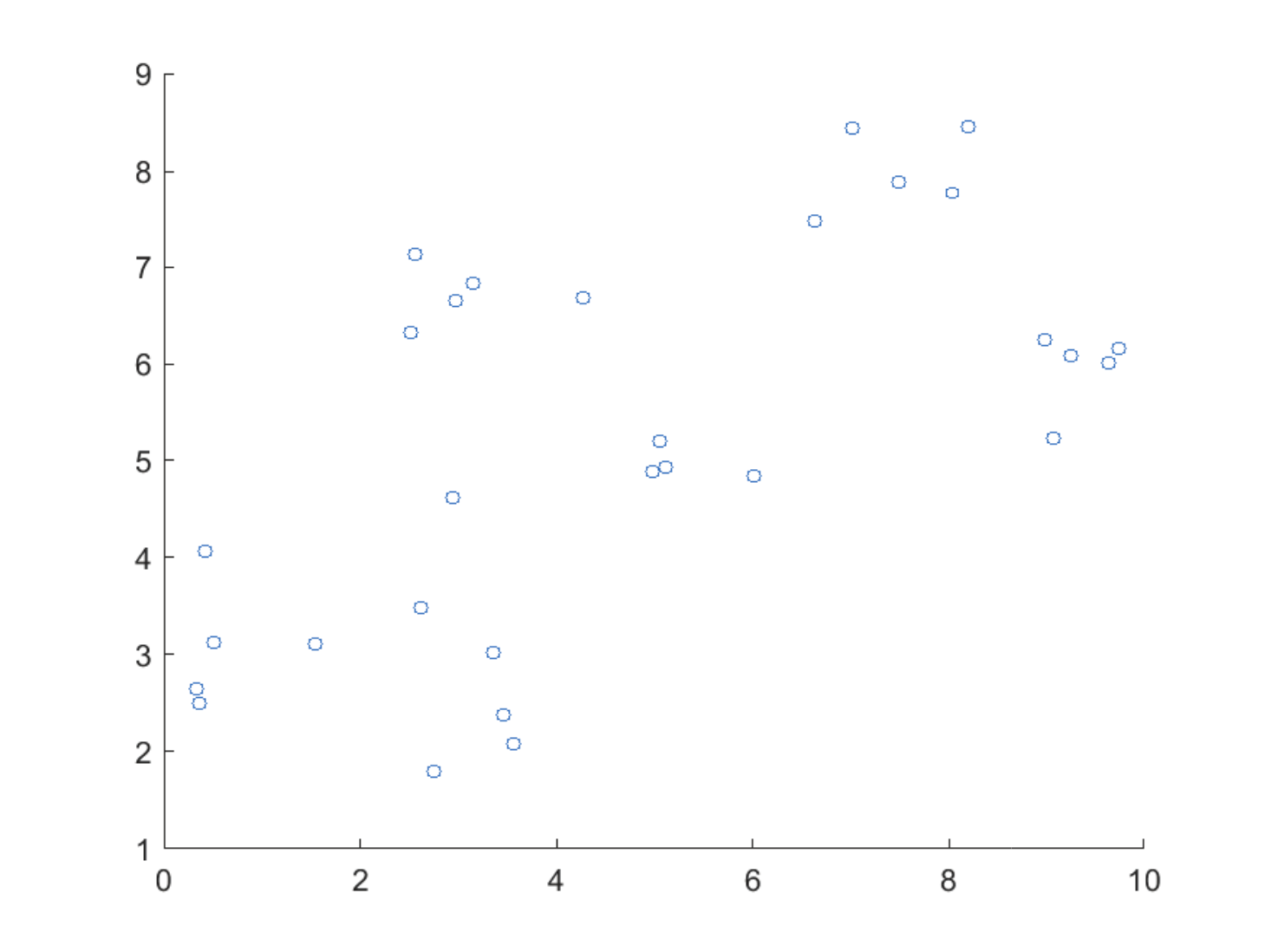}
\caption{30 data points from Gaussian mixture distribution.}
\label{GaussianMixture30}
\end{figure}

\begin{figure}[!t]
\centering
\includegraphics[width=8cm]{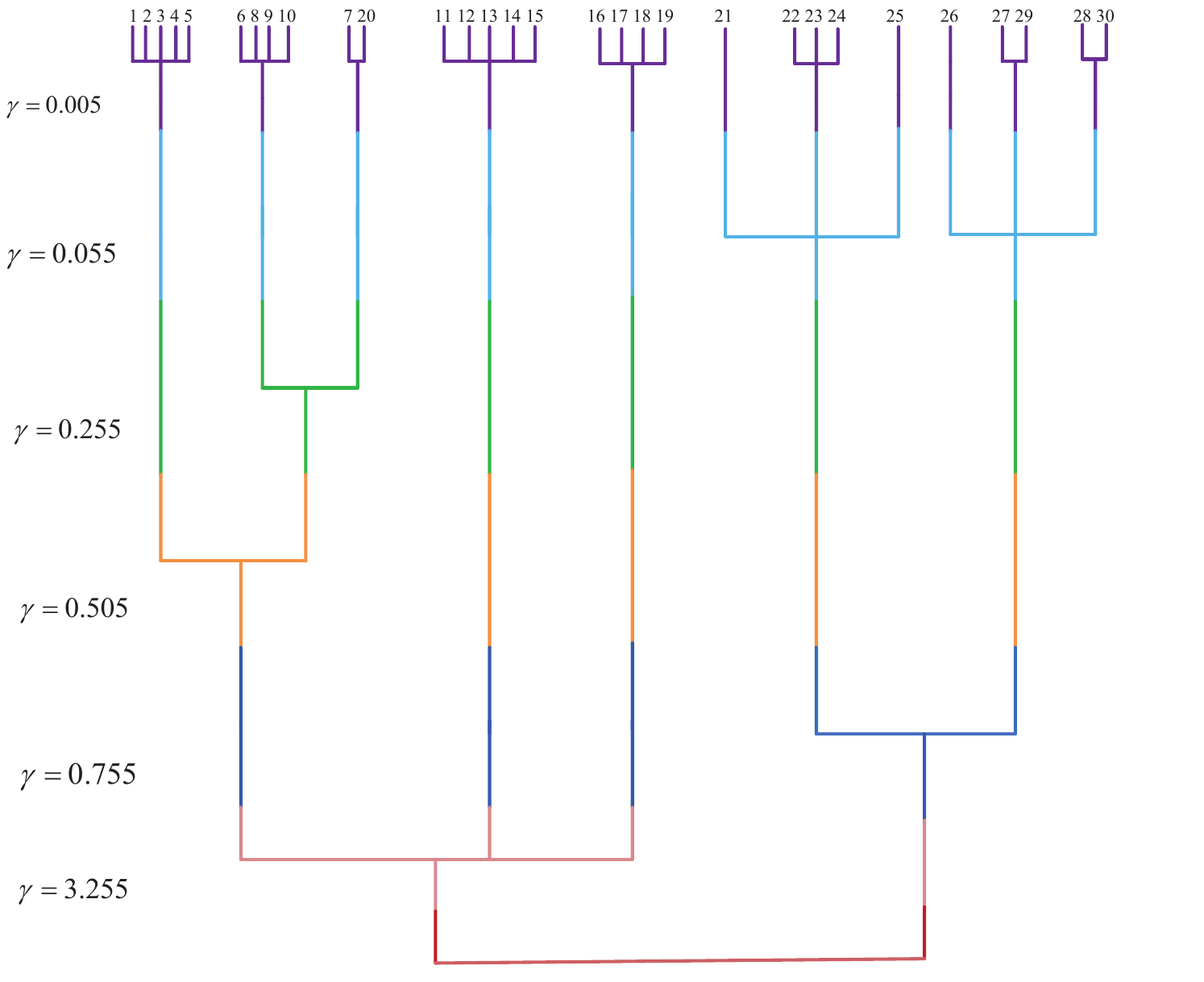}
\caption{Hierarchical cluster path of GaussianMixture30 data obtained by CAF-HFCM.}
\label{GaussianMixture30ClusterPath}
\end{figure}

\subsection{Hierarchical Cluster Path Comparison}
To further demonstrate the fact that CAF-HFCM is capable of yielding a hierarchy while FCM does not have such ability, we perform both CAF-HFCM and FCM over a set of cluster numbers on GaussianMixture data set and the respective cluster centroids obtained by the both methods are shown in Figure \ref{CAF-HFCM-GaussianMixtureU} and Figure \ref{FCM-GaussianMixtureU}. For clarity and simplicity of discussion, GaussianMixture data set generated from six-component Gaussian mixture distribution is shown in Figure \ref{GaussianMixShow}, where each color corresponds to a component and the numbers are marks of respective ingredients. As shown in Figure \ref{CAF-HFCM-GaussianMixtureU}, for CAF-HFCM, as the number of clusters drops from 19 (Figure \ref{FCM-U191}) to 18 (Figure \ref{FCM-U181}), two cluster centroids in component 3 are automatically fused, while data points belonging to the two centroids naturally agglomerate together. Similarly, as the number of clusters reduces from 18 (Figure \ref{FCM-U181}) to 17 ({Figure \ref{FCM-U171}}), two cluster centroids in component 4 are automatically merged, while data points belonging to the two centroids naturally agglomerate together. In contrast, for FCM, as the cluster number is from 19 (Figure \ref{FCM-U19}) to 18 (Figure \ref{FCM-U18}), the number of centroids in component 1 raises from 2 to 3 while that in component 2 changes from 3 to 4, which means that some data points belonging to one cluster as K = 19 are not grouped into similar cluster any more as K = 18. Similarly, the phenomenon happens again when K is from 18 (Figure \ref{FCM-U18}) to 17 (Figure \ref{FCM-U17}), i.e., the number of centroids in component 4 raises from 4 to 5. This further confirms the truth that FCM can not construct a hierarchy as traversing every integer from large number to 2. Besides, it is worthy noting that CAF-HFCM only needs once initialization from large cluster number to 2, while {FCM} needs re-initialization for different cluster numbers.

\begin{figure}
\centering
\includegraphics[width=7.5cm]{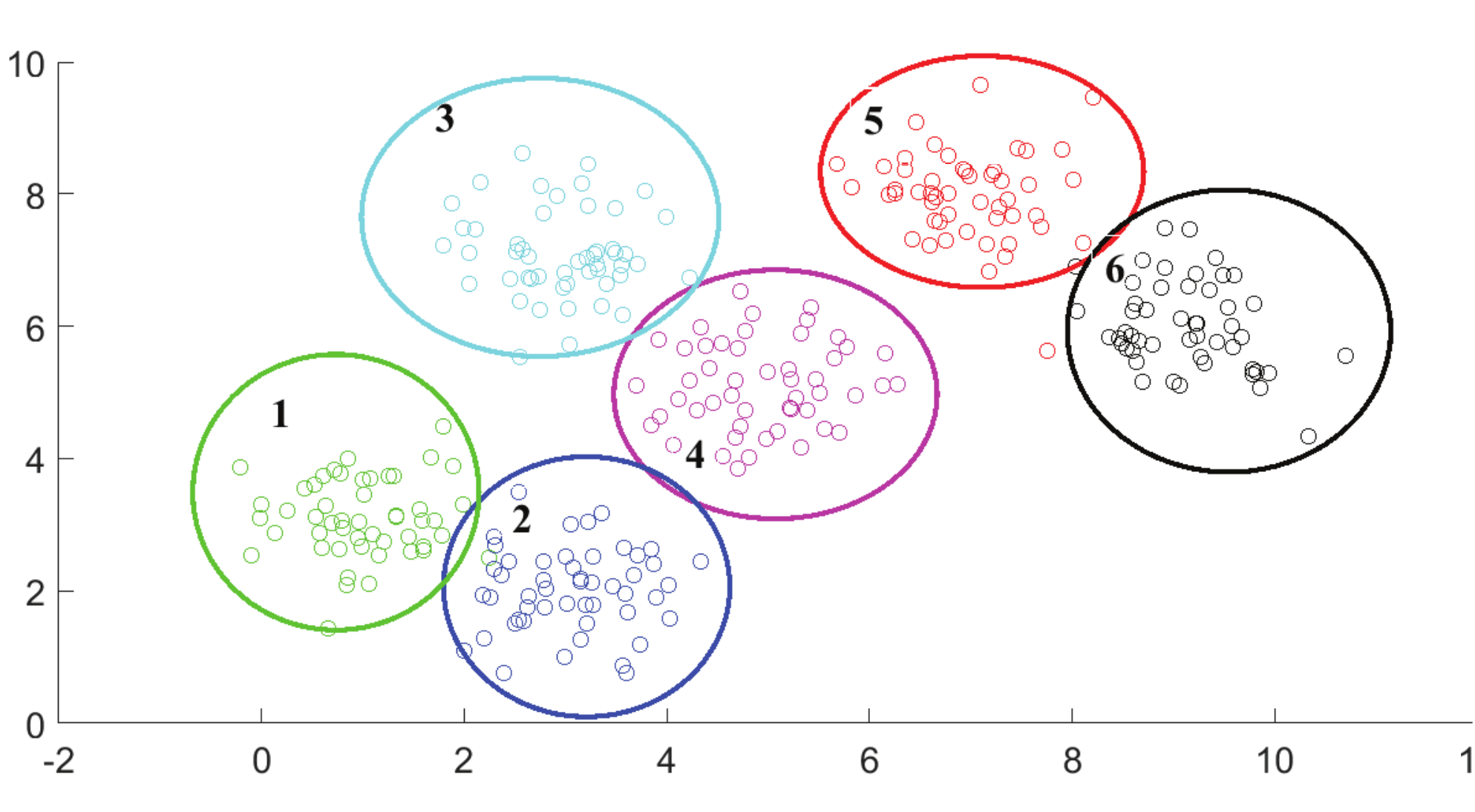}
\caption{GaussianMixture data generated from six-component Gaussian mixture distribution, where each color corresponds to a component and the numbers are marks of respective ingredients.}
\label{GaussianMixShow}
\end{figure}

\begin{figure}[!t]
\centering
\subfigure[$\gamma=0.05 \ (c = 19)$]{%
\includegraphics[width=4.05cm]{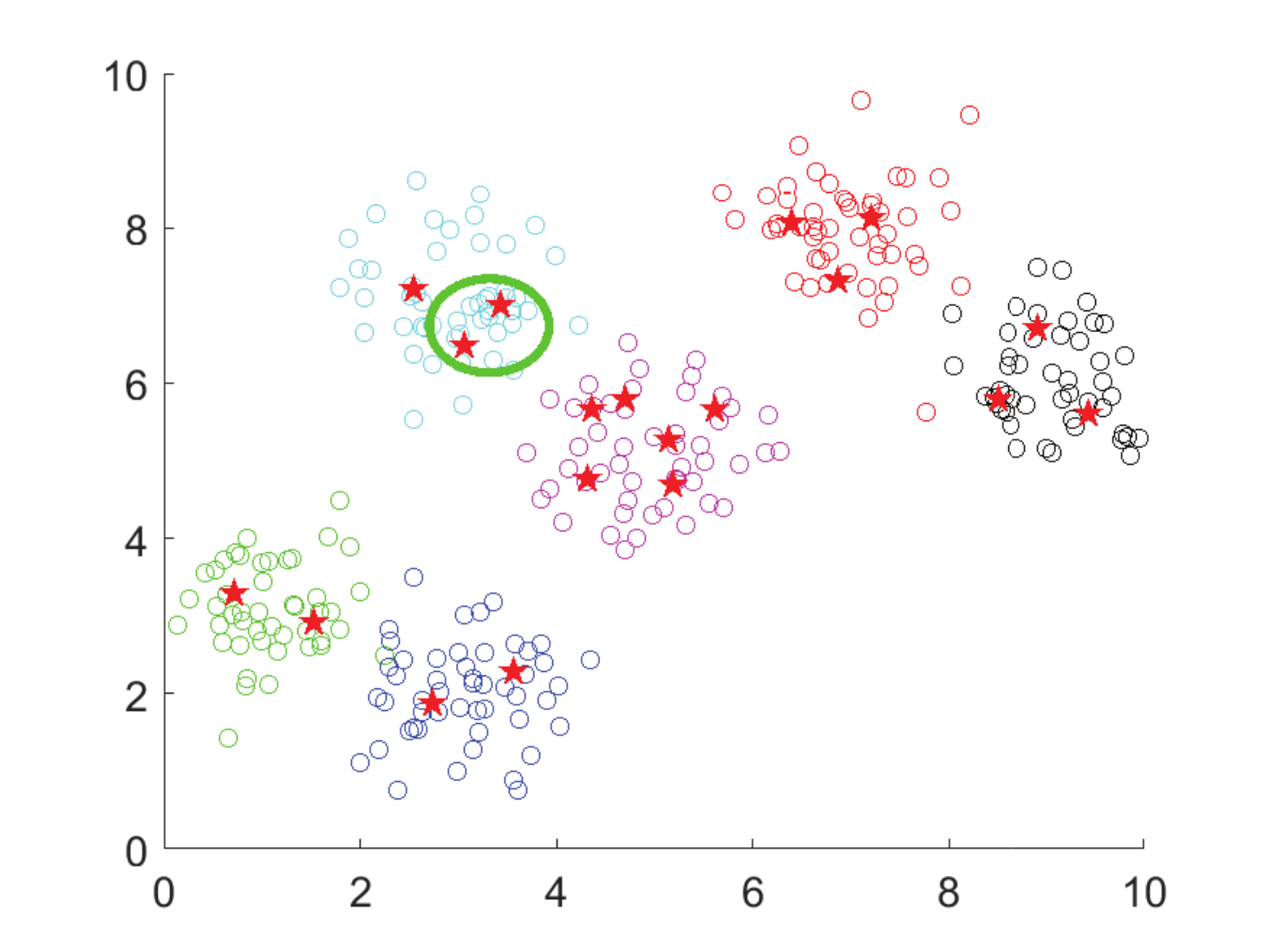}
\label{FCM-U191}}
\quad
\subfigure[$\gamma=0.065 \ (c = 18)$]{%
\includegraphics[width=4.05cm]{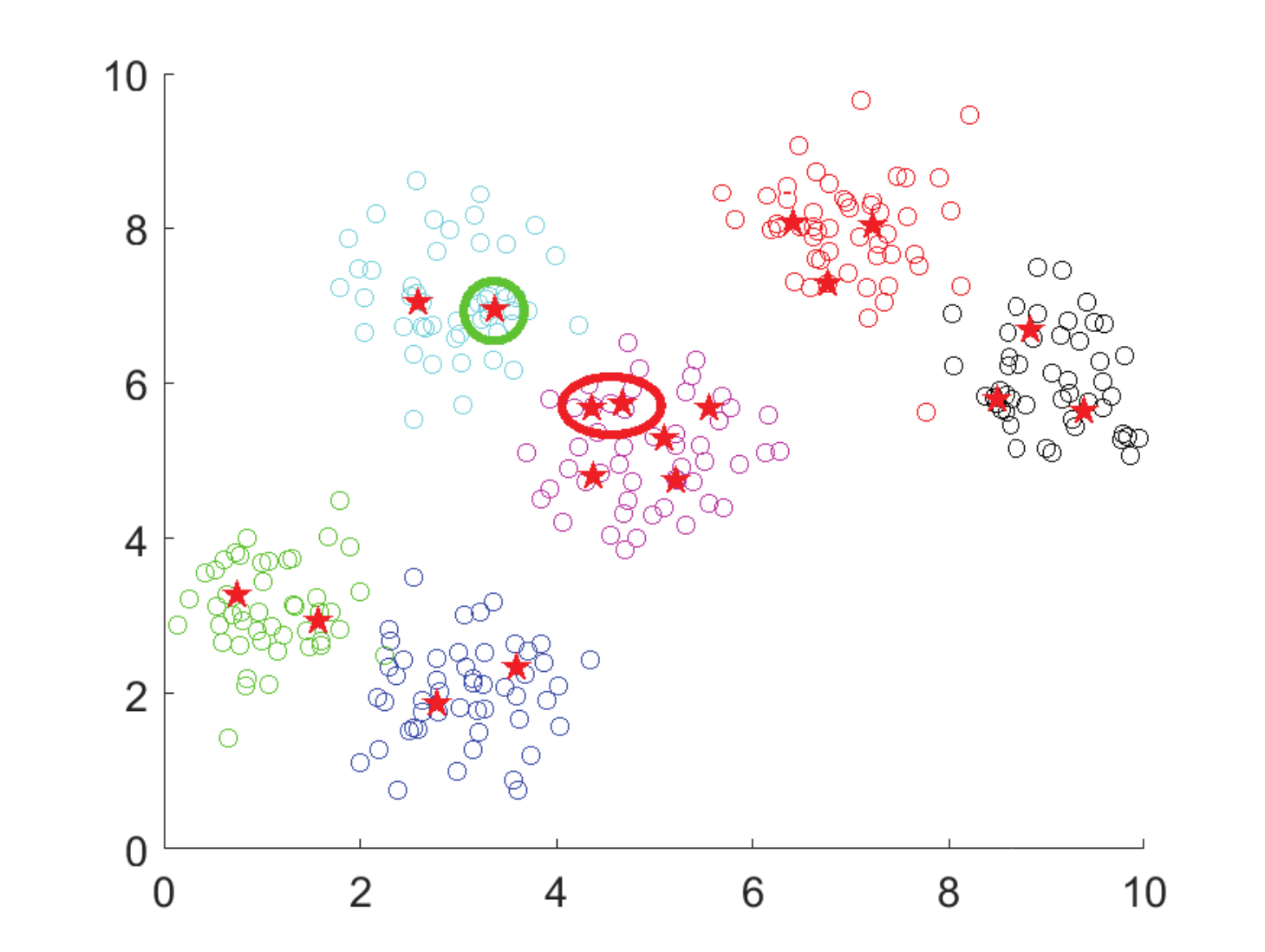}
\label{FCM-U181}}

\subfigure[$\gamma=0.08 \ (c = 17)$]{%
\includegraphics[width=4.05cm]{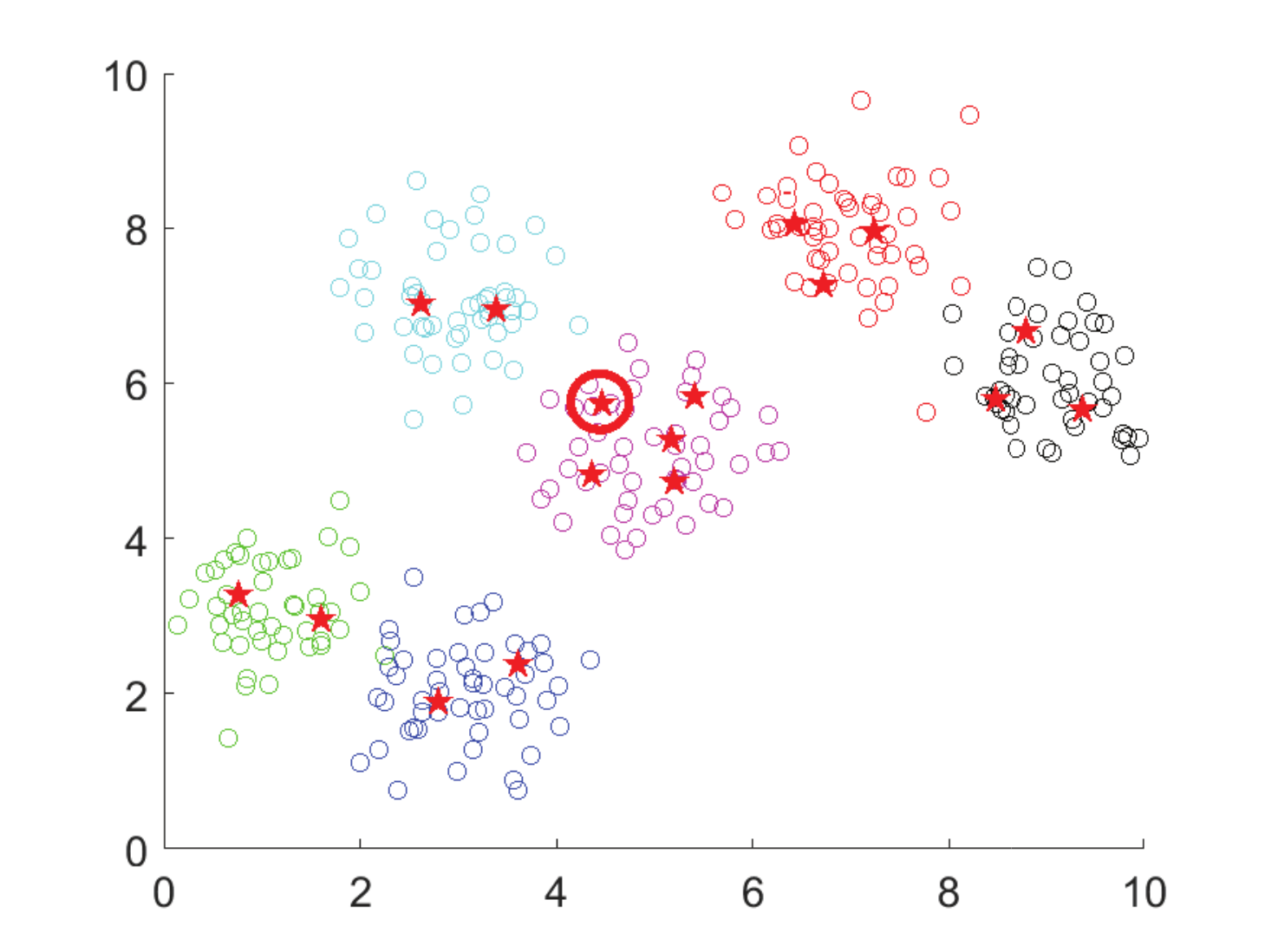}
\label{FCM-U171}}
\quad
\subfigure[$\gamma=0.135 \ (c = 14)$]{%
\includegraphics[width=4.05cm]{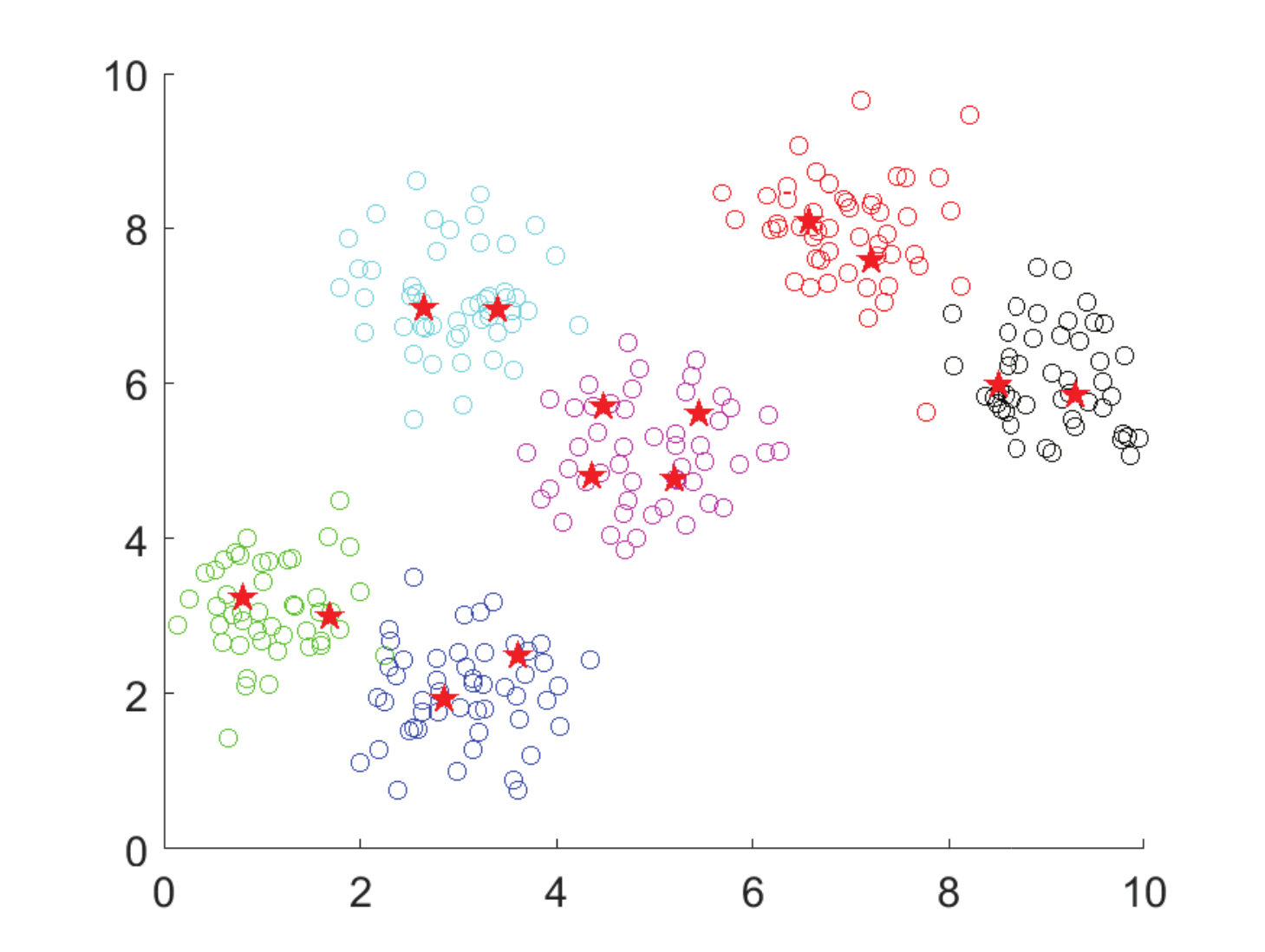}
\label{FCM-U14}}

\subfigure[$\gamma=0.38 \ (c = 9)$]{%
\includegraphics[width=4.05cm]{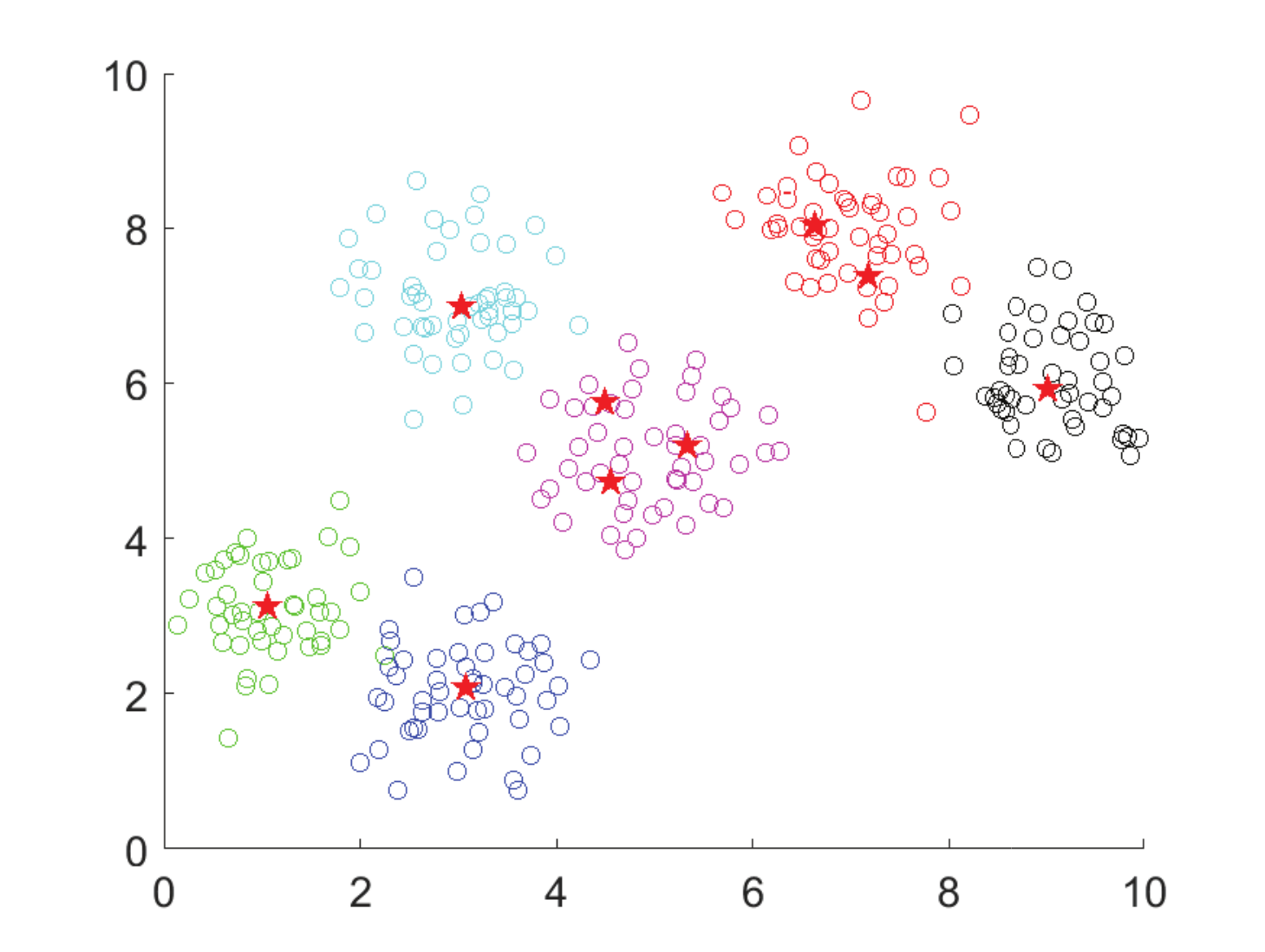}
\label{FCM-U9}}
\quad
\subfigure[$\gamma=0.545 \ (c = 6)$]{%
\includegraphics[width=4.05cm]{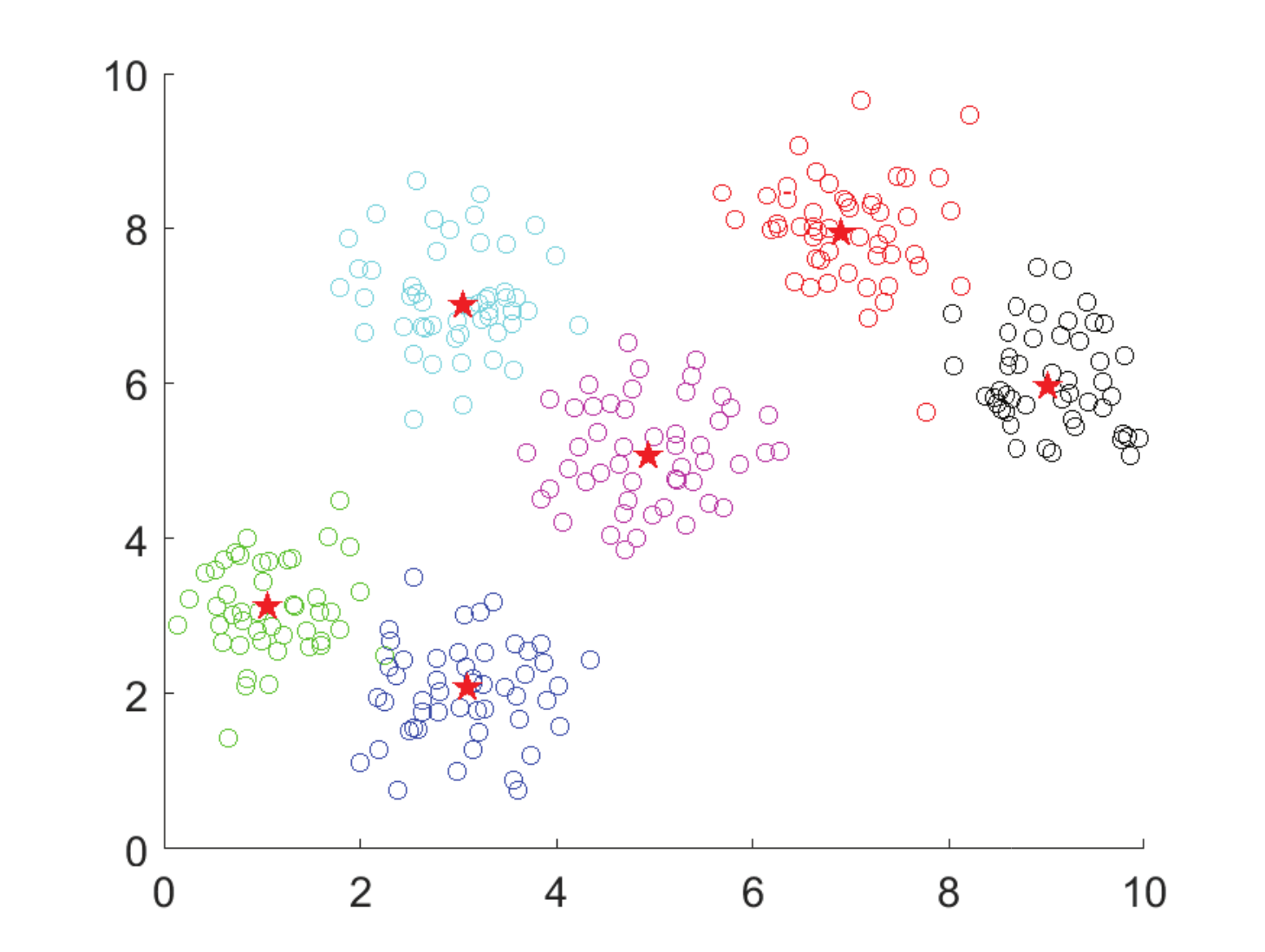}
\label{FCM-U6}}
\caption{Centroids obtained by CAF-HFCM over different cluster numbers.}
\label{CAF-HFCM-GaussianMixtureU}
\end{figure}

\begin{figure}[!t]
\centering
\subfigure[K = 19]{%
\includegraphics[width=4.05cm]{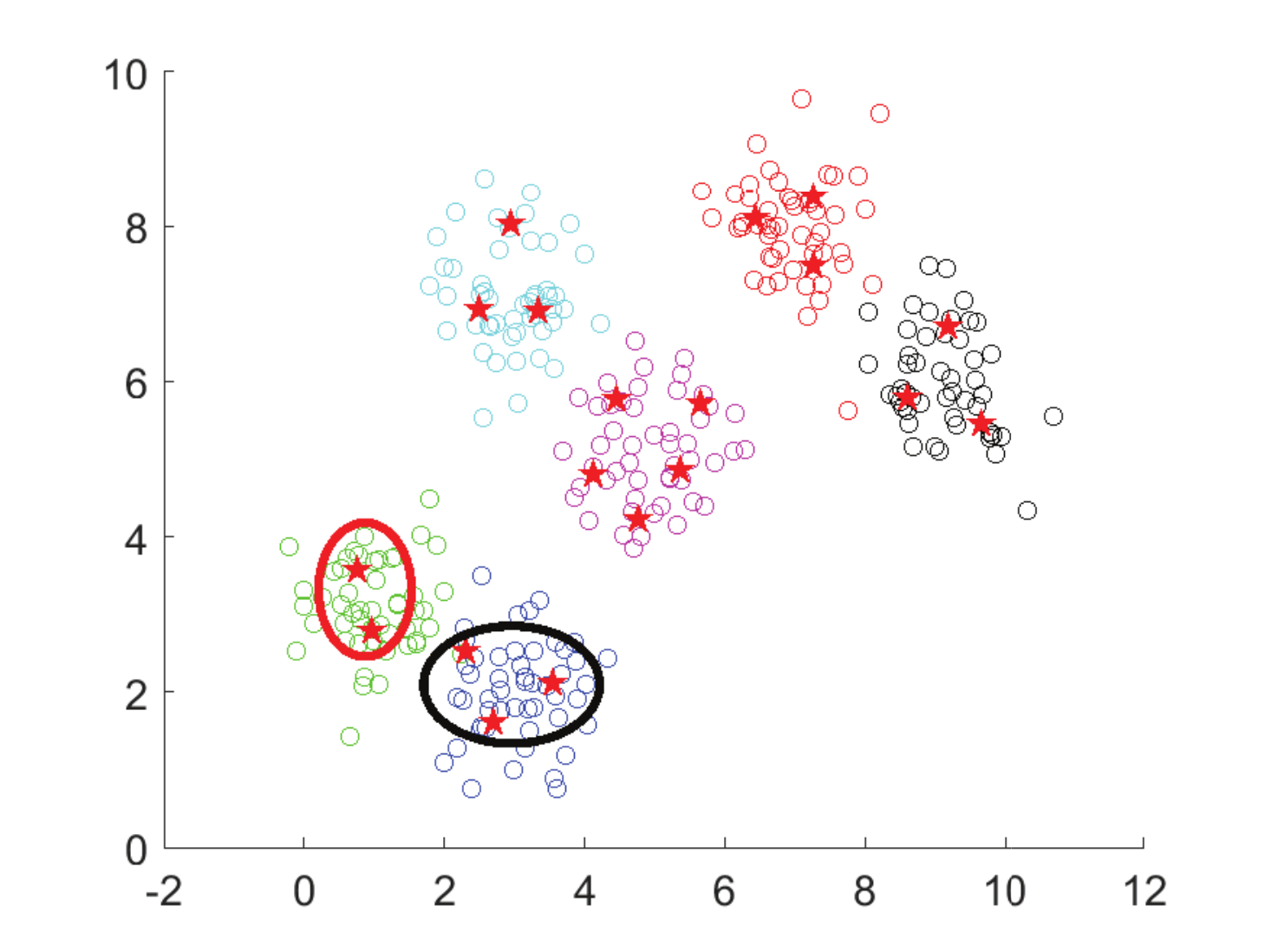}
\label{FCM-U19}}
\quad
\subfigure[K = 18]{%
\includegraphics[width=4.05cm]{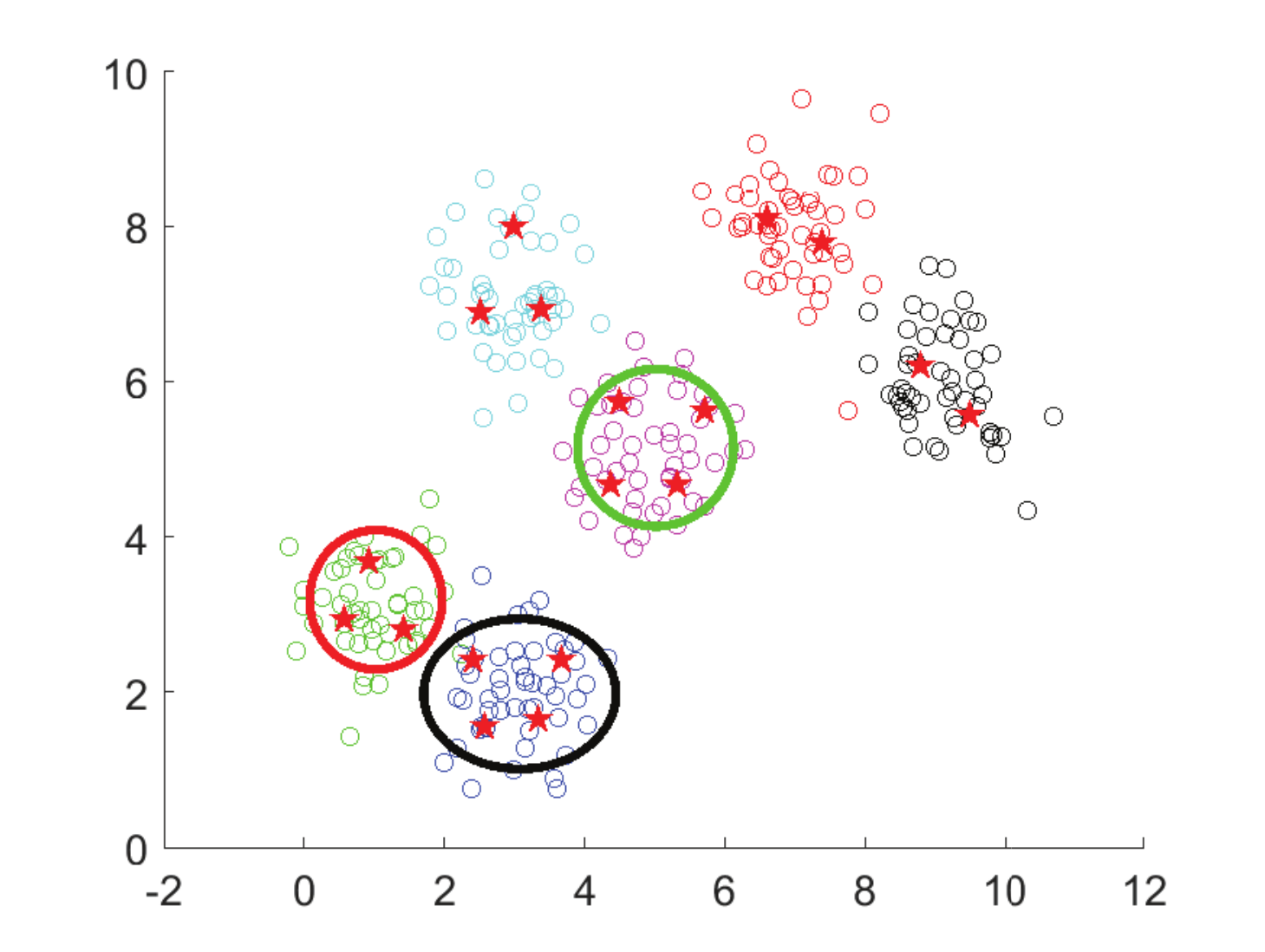}
\label{FCM-U18}}

\subfigure[K = 17]{%
\includegraphics[width=4.05cm]{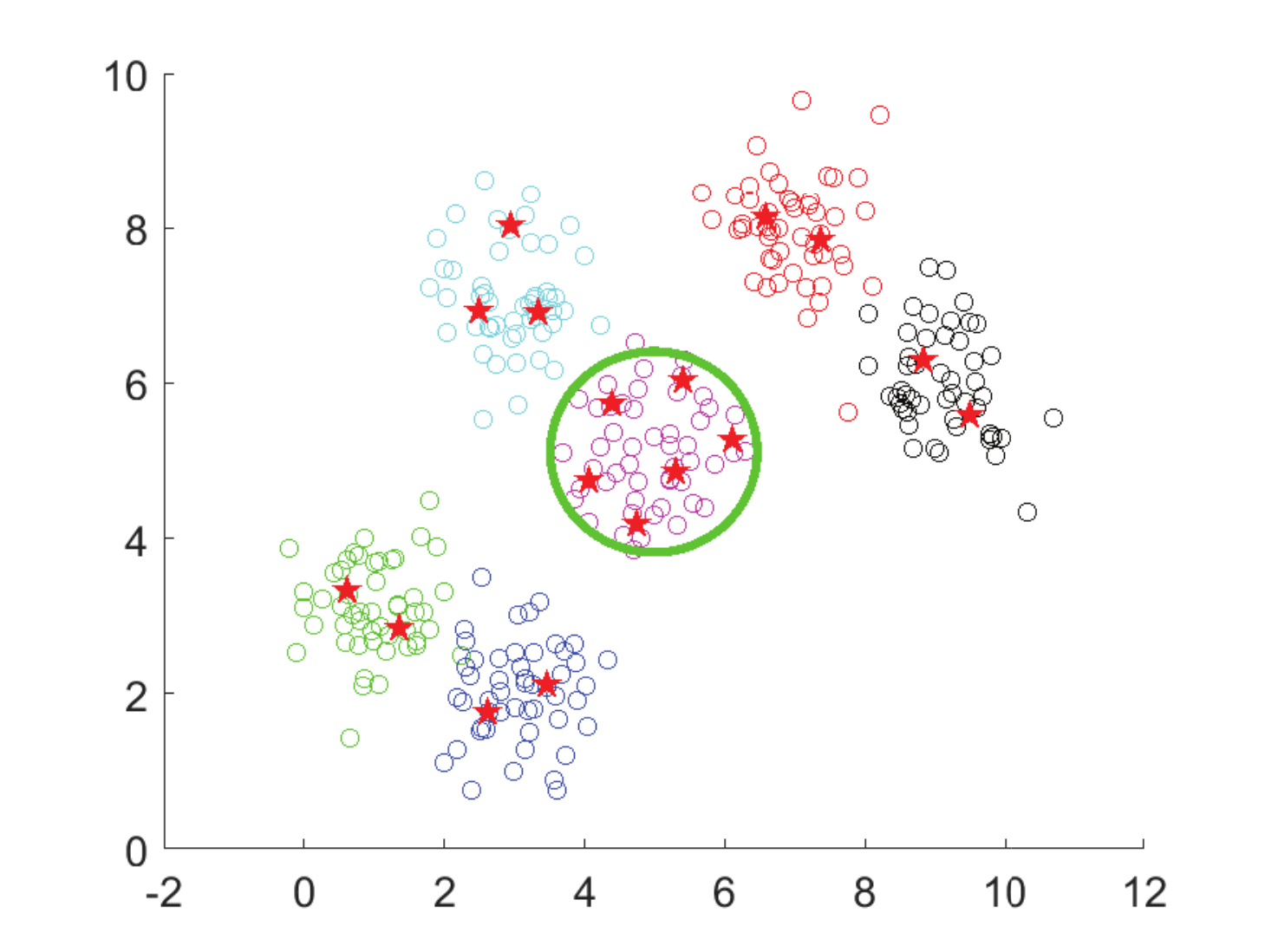}
\label{FCM-U17}}
\quad
\subfigure[K = 14]{%
\includegraphics[width=4.05cm]{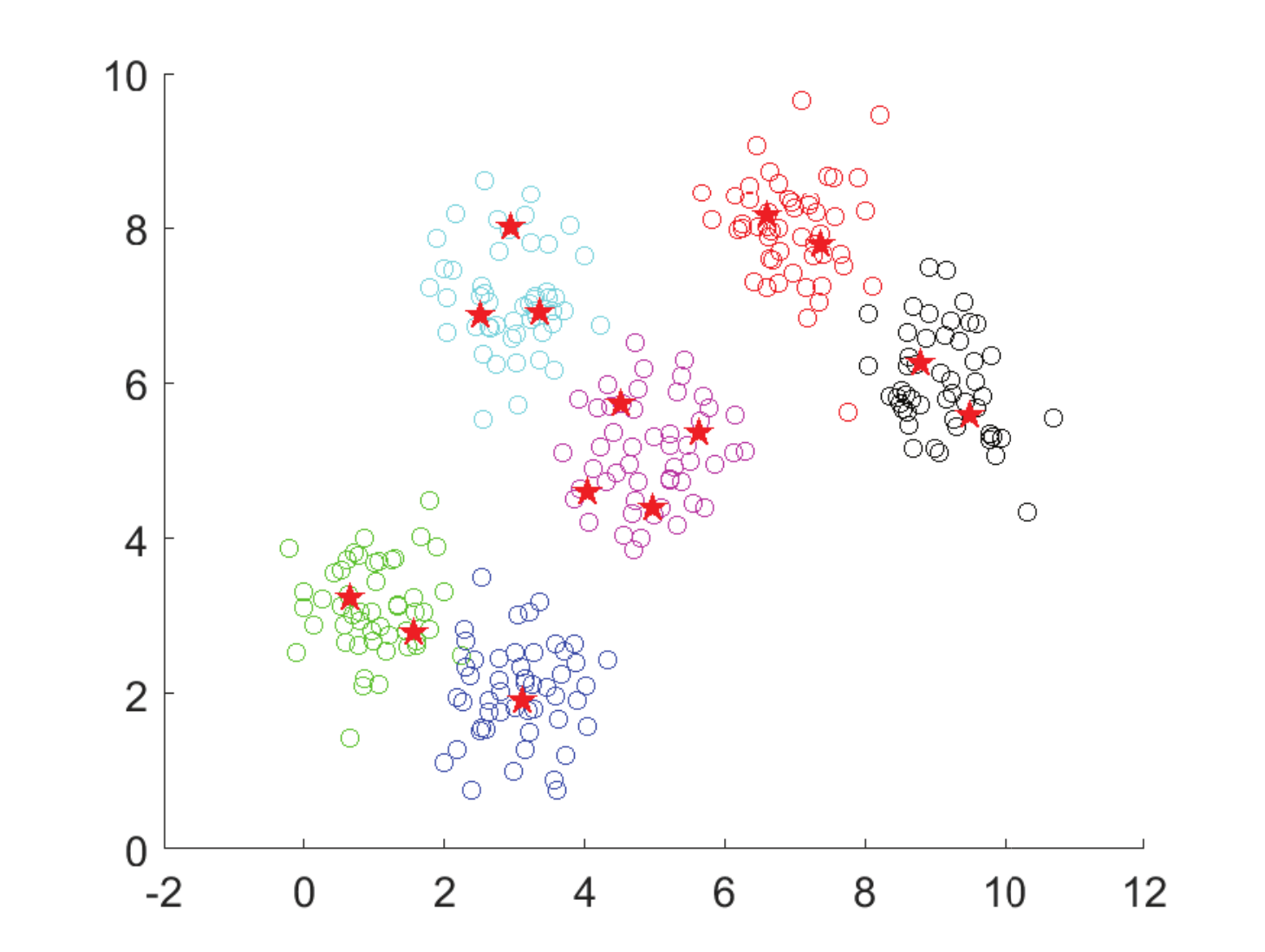}
\label{FCM-U14}}

\subfigure[K = 9]{%
\includegraphics[width=4.05cm]{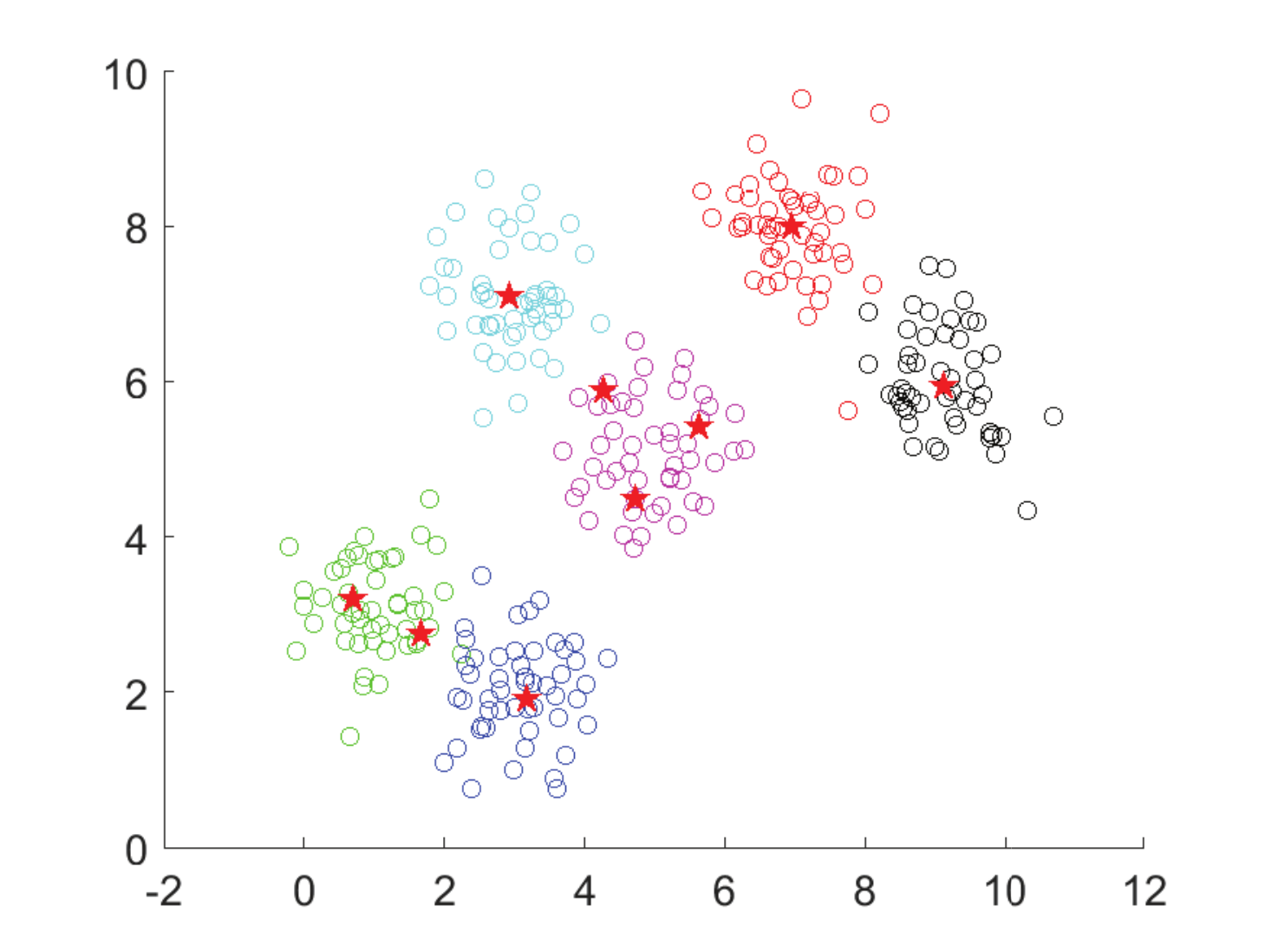}
\label{FCM-U9}}
\quad
\subfigure[K = 6]{%
\includegraphics[width=4.05cm]{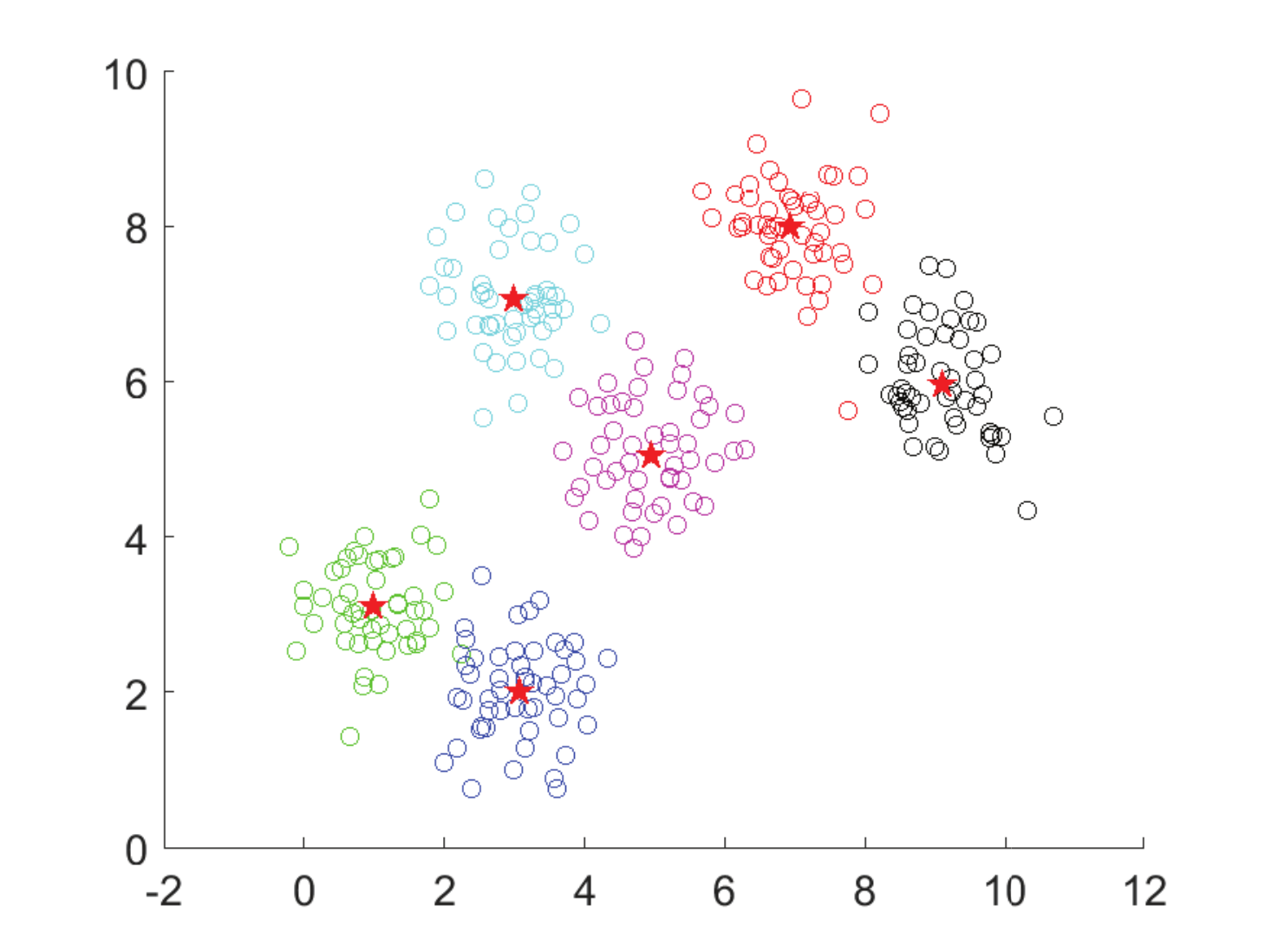}
\label{FCM-U6}}
\caption{Centroids obtained by FCM over different cluster numbers.}
\label{FCM-GaussianMixtureU}
\end{figure}

\subsection{Cluster Index Comparison}
  In this subsection, we {aim at} test{ing} the change trends of different cluster indices with respect to different cluster number. {Seeds dataset is next chosen.} In Figure \ref{SeedsIndex}, the red curve, blue dash line, the green line {indicate} RI, ARI, NMI, separately. From Figure \ref{SeedsIndex}, the highest values 0.8814, 0.7331, 0.7229 are achieved {over} the {optimal} number of clusters, therefore we can conclude that the optimal partitions can be obtained under the optimal size of clusters.

\begin{figure}
\centering
\includegraphics[width=8cm]{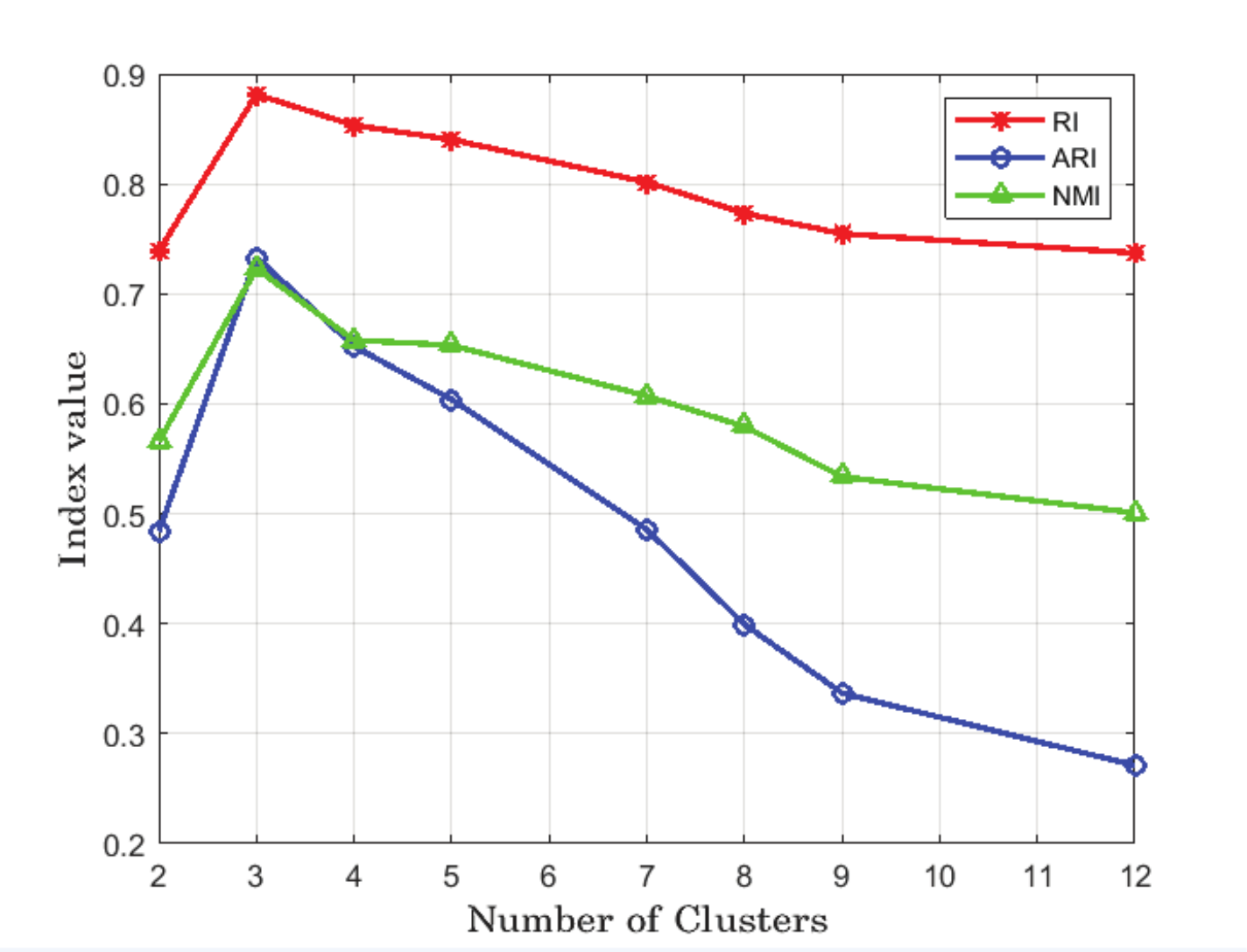}
\caption{Index value obtained by CAF-HFCM performing on Seeds dataset under the sequence of cluster number.}
\label{SeedsIndex}
\end{figure}

\subsection{Robustness Test}

  In the robustness test, another data set covering 13 blocks formed from continuously uniform distribution is generated, where each block has 50 data points, then we add 100 noisy points in the Uniform data, and the noisy Uniform data is shown in {Figure \ref{NoisyUniformData}}. We run CAF-HFCM with $\gamma = 0.55 \ (c = 13)$ and FCM with $c = 13$ on the noisy data set for 20 times, and the corresponding clustering results are shown in {Table \ref{Robustness}}. From the {Table \ref{Robustness}}, we can find that CAF-HFCM performs higher 0.76$\%$ in RI, 5.19$\%$ in ARI, 1.96$\%$ in NMI than FCM. Accordingly, we can conclude that CAF-HFCM is more robust compared with FCM. Furthermore, the variance value with 0 of the clustering results obtained by CAF-HFCM illustrates that CAF-HFCM is less sensitive to initialization than FCM once more. Moreover, further to illustrate that CAF-HFCM performs better than FCM when clustering noisy data, {Figure \ref{NoisyUniformData-CAF-HFCM}} corresponding to CAF-HFCM with $\gamma = 0.55$ and {Figure {\ref{NoisyUniformData-FCM}}} corresponding to FCM with $c = 13$ are shown. Comparing Figure {\ref{NoisyUniformData-CAF-HFCM}} and Figure {\ref{NoisyUniformData-FCM}}, we can see that even FCM is given the optimal cluster number, the optimal partition is not always determined. Different from FCM, the {consistently} optimal partition {always can be obtained} by CAF-HFCM with the corresponding hyperparameter.

\begin{figure}[tbp]
\centering
\includegraphics[height=7.2cm,width=8cm]{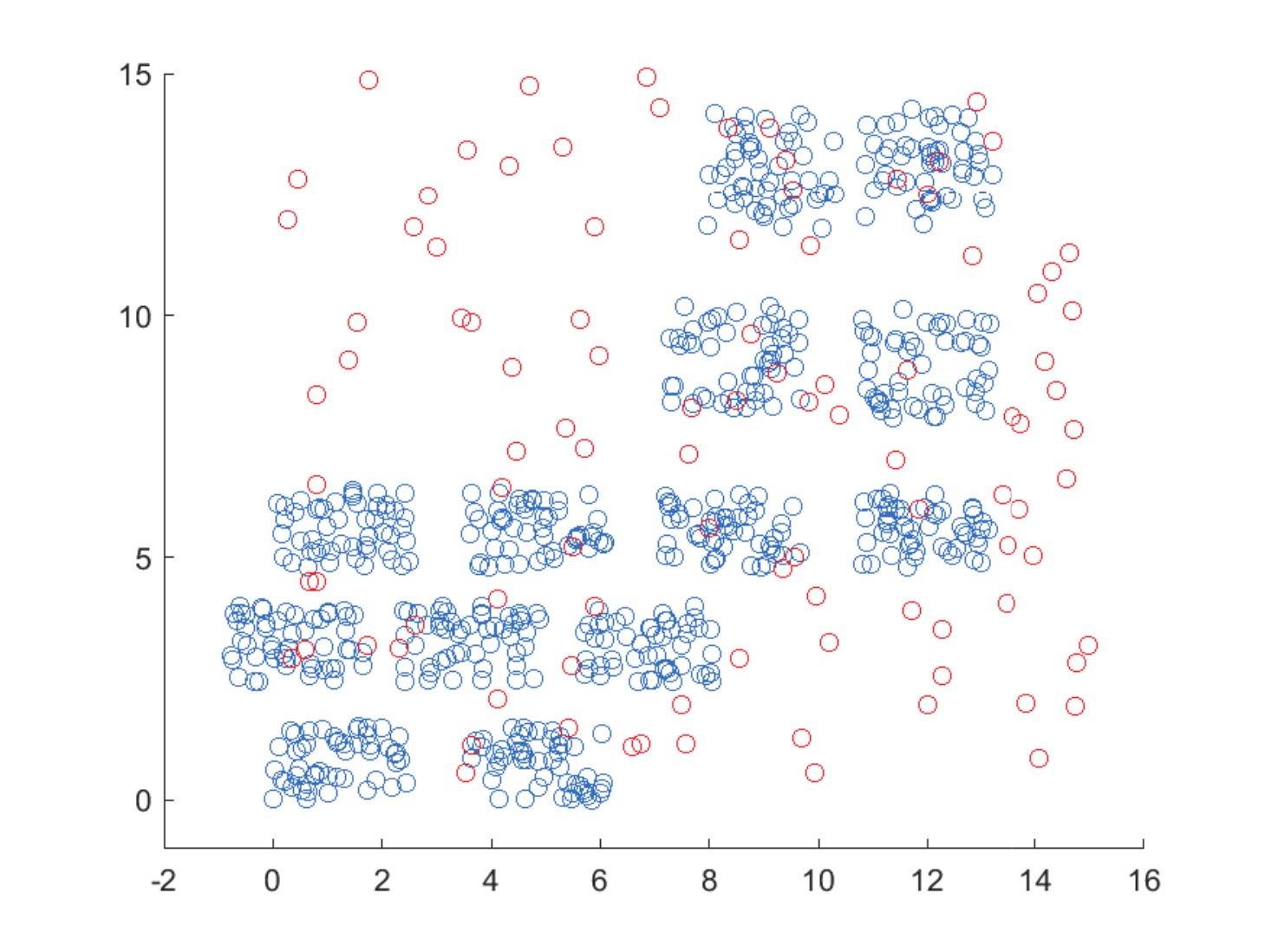}
\caption{Uniform data adding 100 noisy points, where the blue circle and the red circle {indicate} the uniform data and noisy data respectively.}
\label{NoisyUniformData}
\end{figure}

\begin{table}[htbp]
\centering
\caption{RI, ARI, and NMI from CAF-HFCM and FCM performing on the noisy Uniform data with the true number $c$ of clusters.}
\begin{tabular}{cccccccccccc}
\hline
Dataset & Index & CAF-HFCM & FCM
\\ \hline
\multirow{3}{*}{K = 13} & RI & {\bf 0.9963$\pm$0} & 0.9887$\pm$0.0091\\ 
& ARI & {\bf0.9736$\pm$0}  & 0.9217$\pm$0.0630 \\
& NMI & {\bf 0.9804$\pm$0}  & 0.9608$\pm$0.0306\\
\hline
\end{tabular}
\label{Robustness}
\end{table}

\begin{figure}
\centering
\includegraphics[width=7.2cm]{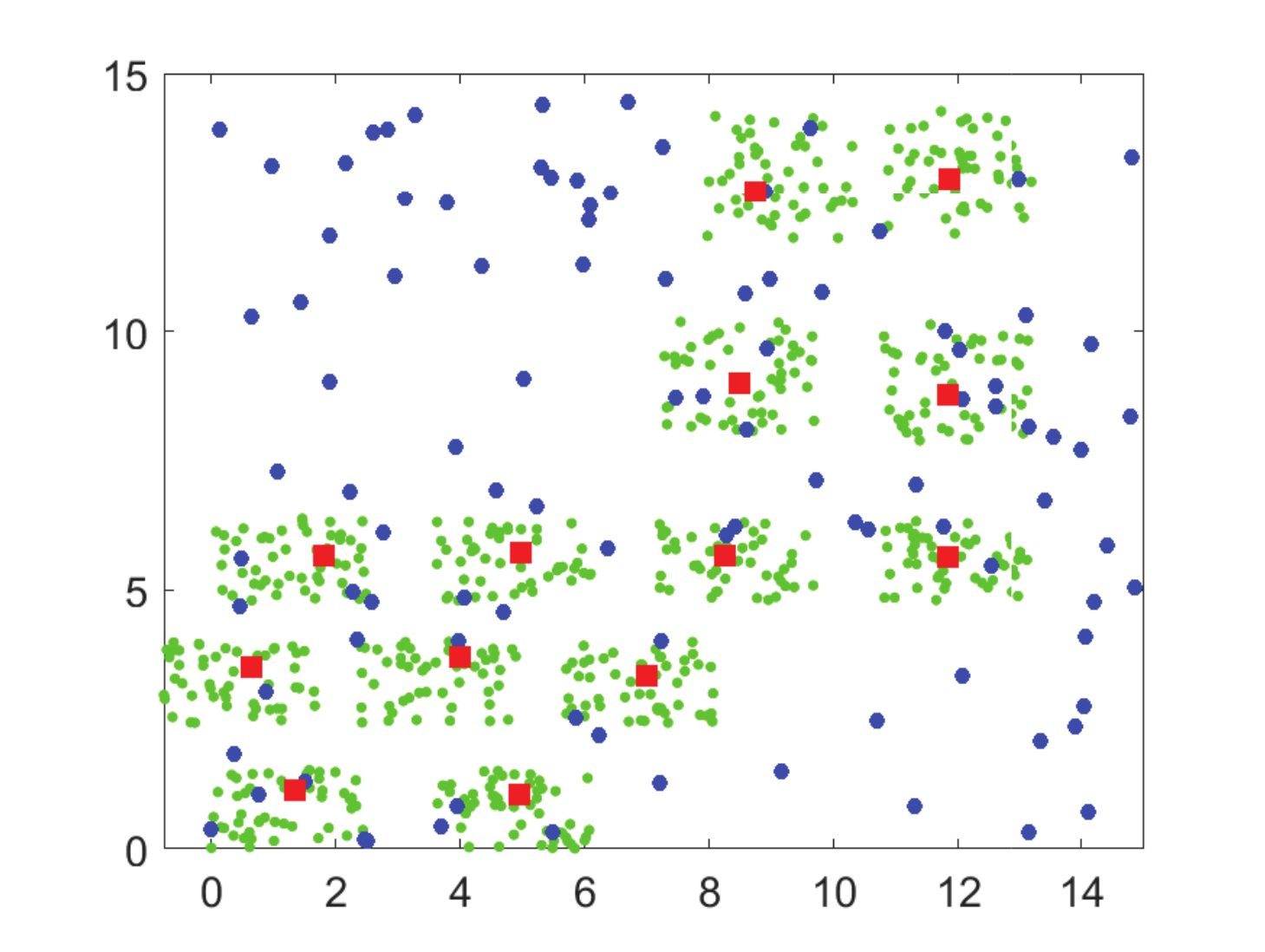}
\caption{Centroids (red squares) of noisy Uniform data obtained by CAF-HFCM with $\gamma=0.55$.}
\label{NoisyUniformData-CAF-HFCM}
\end{figure}

\begin{figure}[htbp]
\centering

\begin{minipage}[t]{1\linewidth}
\centering
\includegraphics[width=7.2cm]{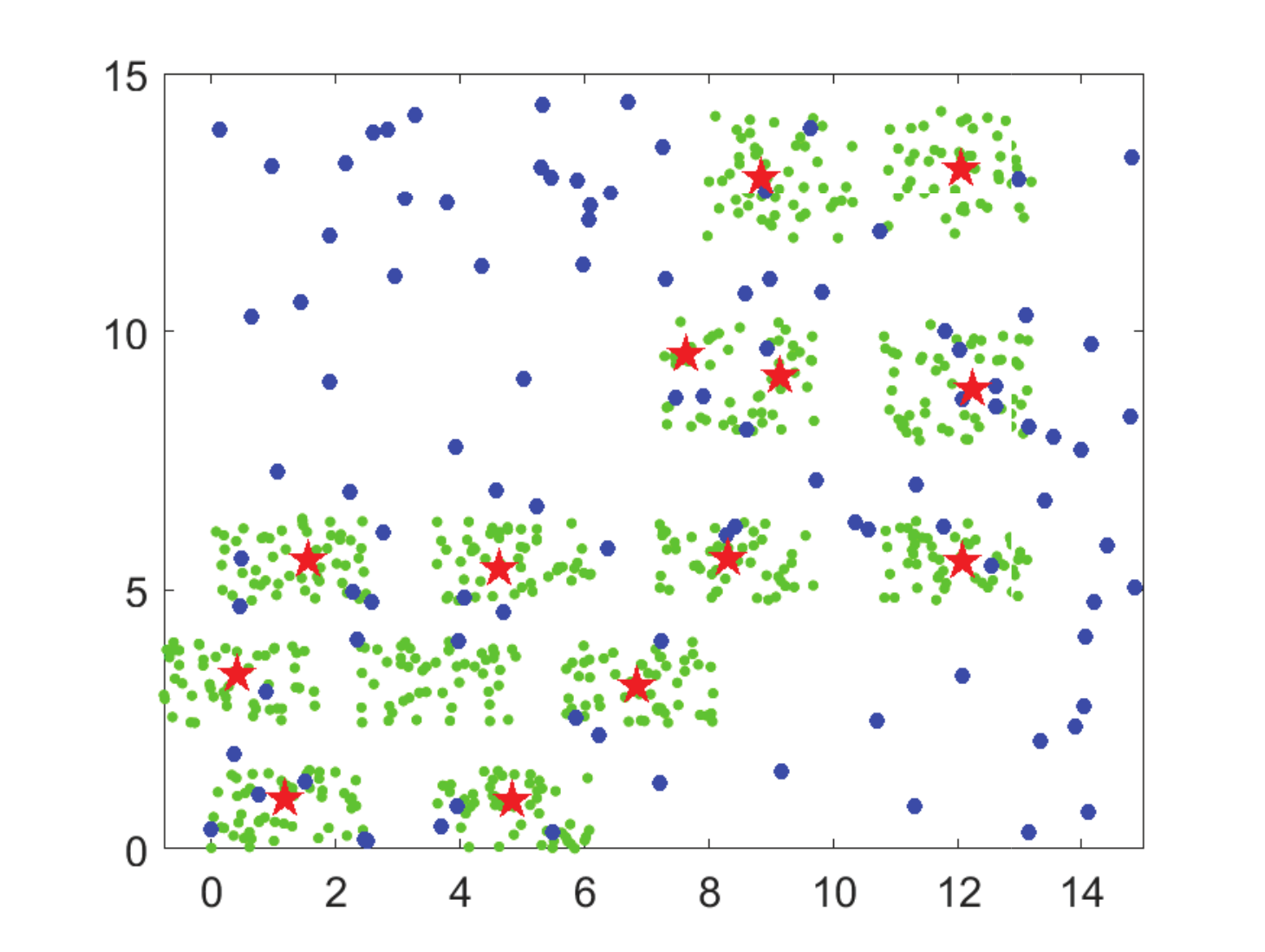}

\includegraphics[width=7.2cm]{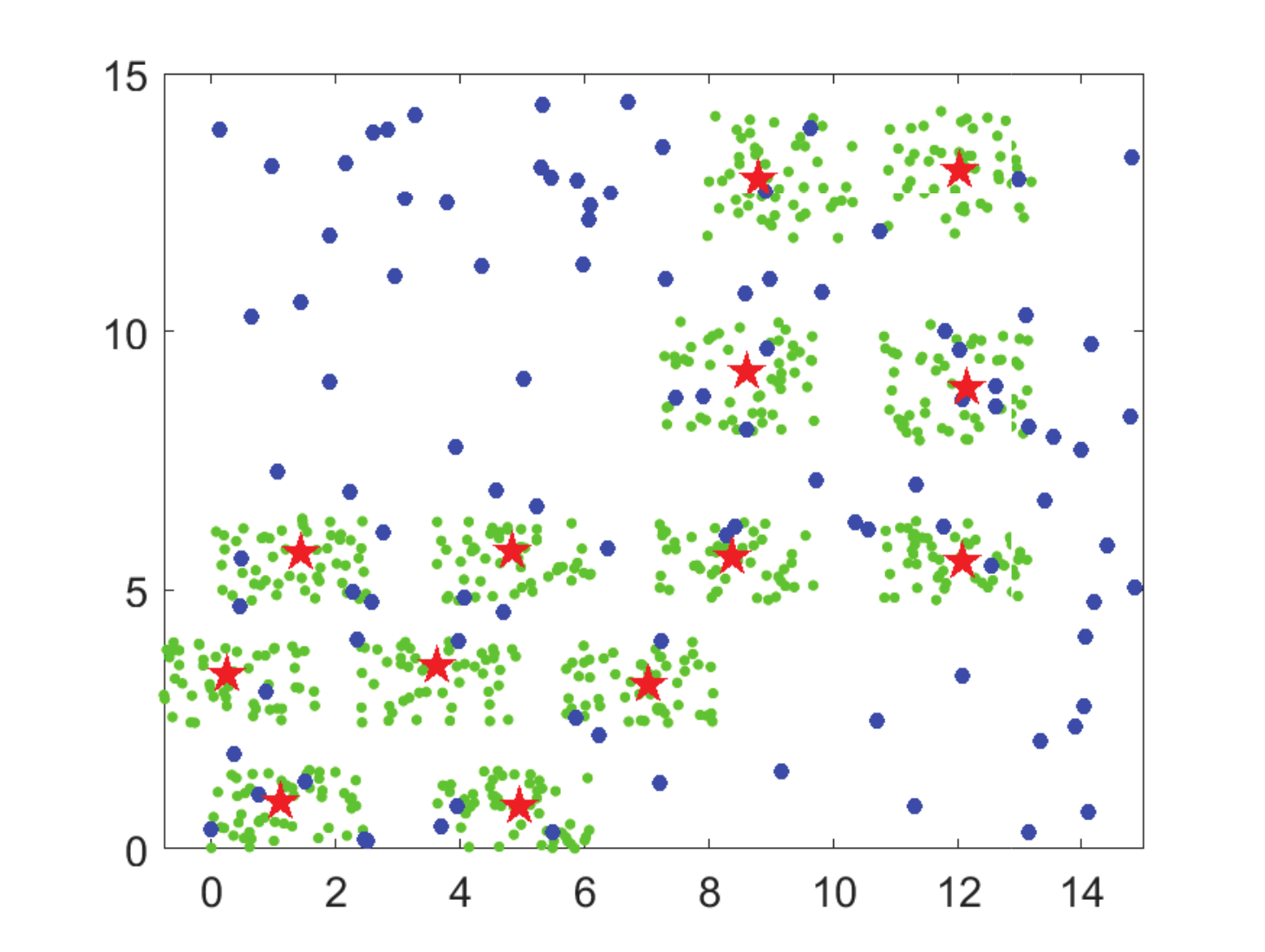}

\includegraphics[width=7.2cm]{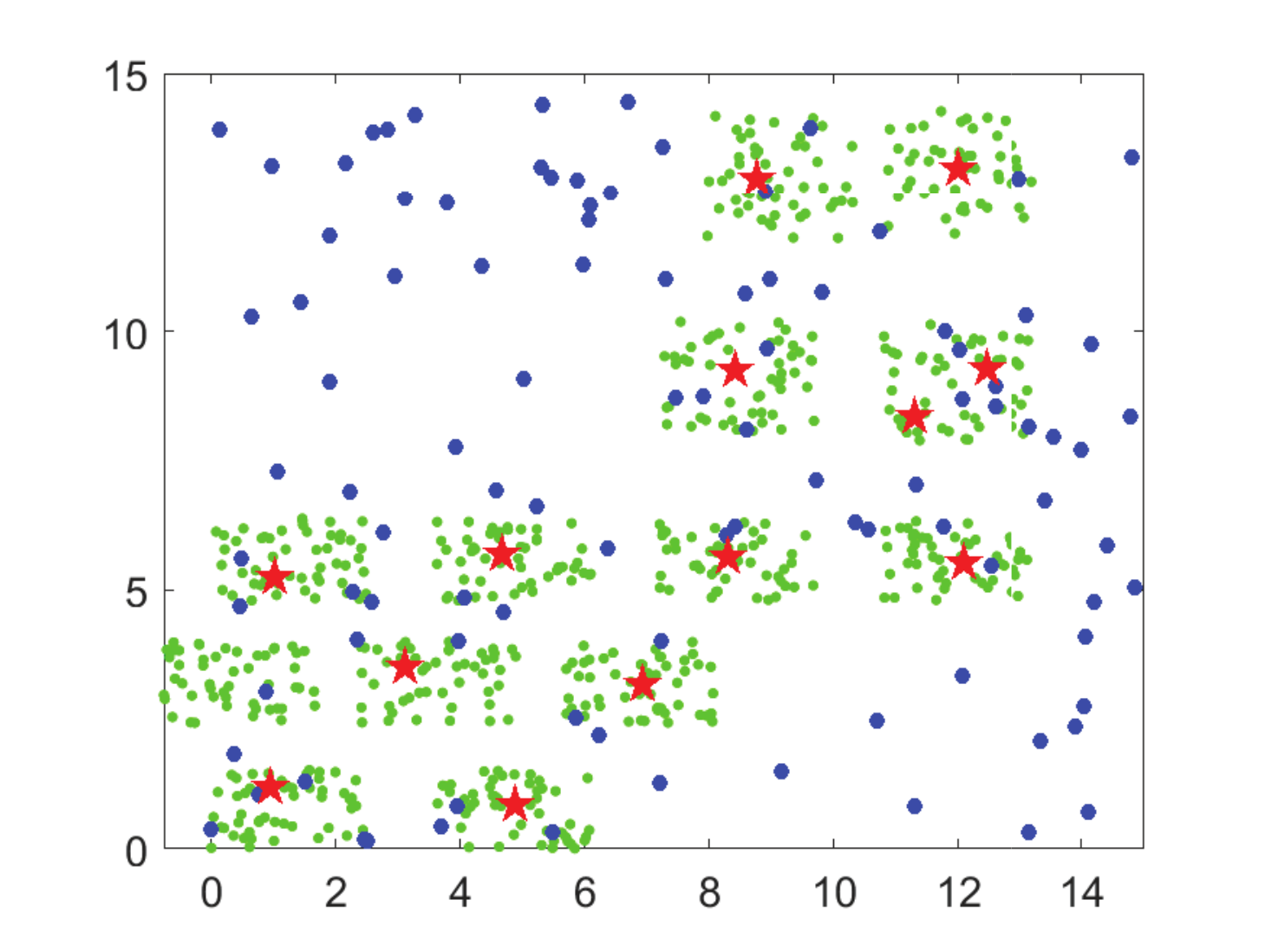}
\end{minipage}%
\centering
\caption{Centroids of noisy Uniform data obtained by FCM over the optimal cluster number $c = 13$.}
\label{NoisyUniformData-FCM}
\end{figure}

\subsection{Convergence Study}

{Without loss of generality, we analyze the convergence of CAF-HFCM on the simulation dataset and real dataset in this section, namely GaussianMixture data and Iris data. For GaussianMixture dataset, we randomly choose 35 data points as the initial centroids, and the 35 initial centroids automatically fuse into 19 centroids over the initial  penalty factor $\gamma$, i.e., GaussianMixture dataset is partitioned into 19 clusters over the initial $\gamma$. Moreover, CAF-HFCM over this initial $\gamma$, i.e., $c=19$ has the largest descend range when it converges among that of all the hyperparameters, so we choose $c = 19$ for GaussianMixture data set to show the values of objective function, and the corresponding convergence curve is shown in Figure \ref{Convergence_GaussianMixtureK19}. Furthermore, from Figure \ref{Convergence_GaussianMixtureK19}, we can see that the algorithm has converged just after 5 iterations even $c$ is relatively large, let alone when $c$ is smaller. Similarly, we choose $c = 8$ for Iris data and the corresponding convergence curve is given in Figure \ref{Convergence_IrisDataK8} and the same fact is found. So as can be seen from both Figure \ref{Convergence_GaussianMixtureK19} and Figure \ref{Convergence_IrisDataK8} that CAF-HFCM has fast convergence rate.}

  \begin{figure}
  \centering
  \includegraphics[width=8.5cm]{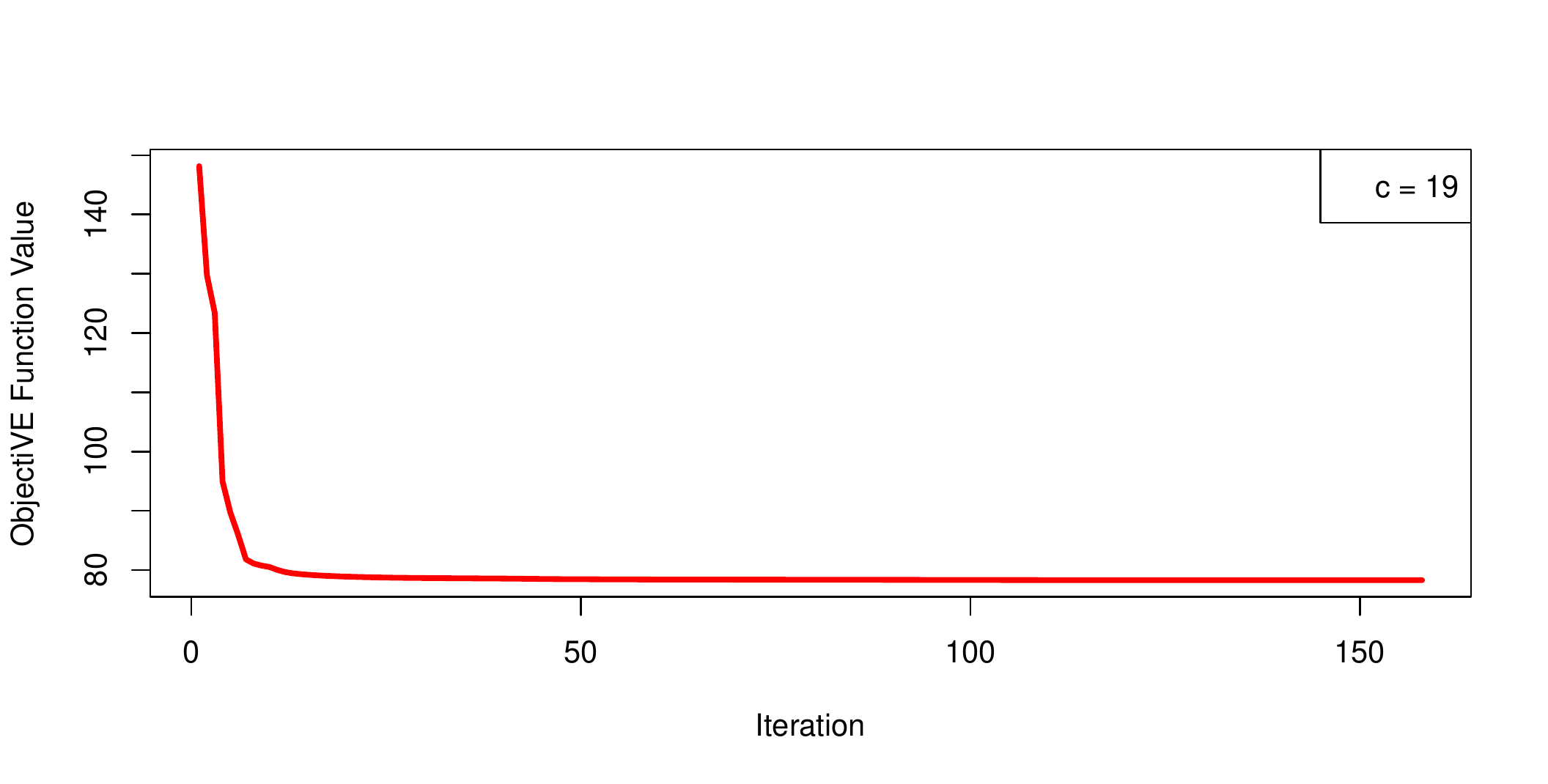}
  \caption{Convergence study on the GaussianMixture data.}
  \label{Convergence_GaussianMixtureK19}
  \end{figure}

  \begin{figure}[!t]
  \centering
  \includegraphics[width=8.5cm]{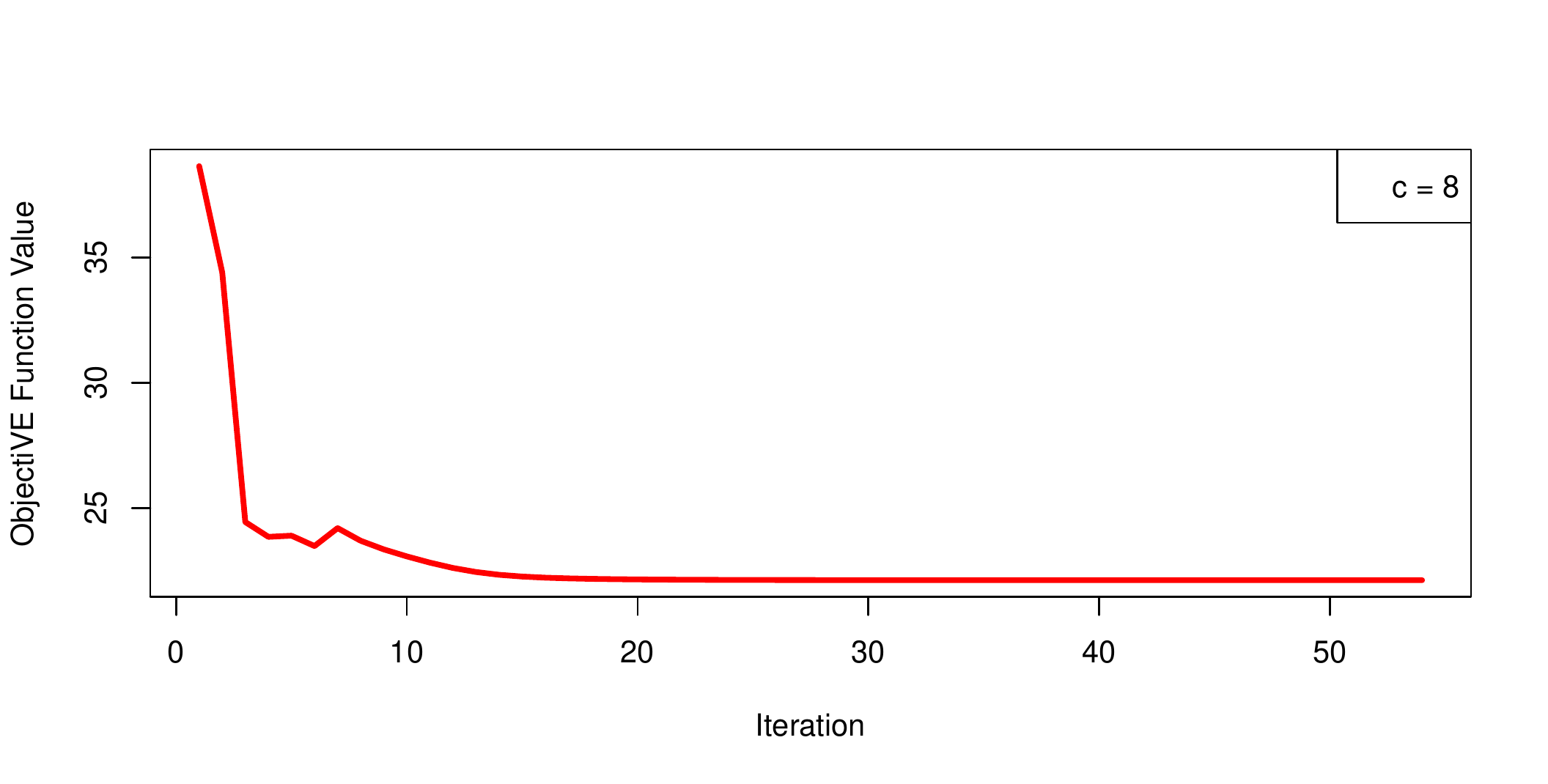}
  \caption{Convergence study on the Iris data.}
  \label{Convergence_IrisDataK8}
  \end{figure}

\section{Conclusion}
In this paper, a Centroid Auto-Fused Hierarchical FCM dubbed as CAF-FCM is proposed, which can not only automatically obtain the optimal number of clusters without resorting to any validity index but also progressively generate a cluster hierarchy as a byproduct in the optimization procedure. Specifically, we randomly choose a relatively large number of data points as the initial centroids, $\ell_{2}$ norm penalty between these initial centriods is next added to FCM's objective to promote the {automatic fusion} of these cluster centroids. Moreover, the corresponding optimization problem is solved by an alternating strategy, in which the subproblem with respect to the membership matrix just contains an analytic solution, the subproblem with respect to the centroid matrix is solved by the ADMM optimization method. The convergence is empirically validated as well. Extensive experiments on both synthetic and real data sets show that CAF-HFCM acquires better performance in comparison with the state-of-art methods. Less sensitivity to initialization and robustness to noisy data points are also observed. Furthermore, CAF-HFCM can straightforwardly be extended to the variants of FCM, such as kernelized fuzzy c-means (KFCM) \cite{ExFCM:D.Q. Zhang}. {However, CAF-HFCM may entail some potential limitations as follows: for very large datasets, $a\sqrt{n}$ is a relatively large number, therefore it is a little slow to update the centroid matrix over small $\gamma$ by using ADMM. For this situation, we can utilize k-means as the preprocess step to obtain some representative data points as the cluster centroids. Besides, as what FCM entails, our CAF-HFCM have some difficulties in dealing with high dimensional datasets. Fortunately, we can resort to the variants of FCM especially proposed for high dimensional datasets.}

\section{Acknowledgments}

This work is supported by  the Key Program of National Natural and Science Foundation of China (NSFC) under Grant No. 61732006 and the NSFC under Grant No. 61672281. We would like to thank Miin-Shen Yang and Yessica Nataliani for providing codes related to paper "Robust-learning fuzzy c-means clustering algorithm with unknown number of clusters".

\ifCLASSOPTIONcaptionsoff
  \newpage
\fi

\end{document}